\documentclass[times,twocolumn,final,authoryear]{elsarticle}

\usepackage{ycviu}
\usepackage{framed,multirow}

\usepackage{amssymb}
\usepackage{latexsym}

\usepackage{url}
\usepackage{xcolor}

\usepackage{times}
\usepackage{epsfig}
\usepackage{graphicx}
\usepackage{amsmath}
\usepackage{amssymb}
\usepackage{fixltx2e}
\usepackage{amsfonts}
\usepackage{listings}
\usepackage{algorithmic}
\usepackage{booktabs} 
\usepackage{multirow}
\usepackage{amssymb} %added it for the L in the caption of Fig. 1
\usepackage{gensymb}
\usepackage{mathtools}
\usepackage{adjustbox}

\usepackage[labelfont=bf]{caption}

\definecolor{DarkGreen}{rgb}{0.2,0.5,0.2} % to color links in references

\makeatletter
\AtBeginDocument{\def\@citecolor{DarkGreen}}
\makeatother

\usepackage[colorlinks,citecolor=DarkGreen,urlcolor=DarkGreen,bookmarks=false,hypertexnames=true]{hyperref} 

\begin{document}

\ifpreprint
  \setcounter{page}{1}
\else
  \setcounter{page}{1}
\fi

\begin{frontmatter}

\title{Facial Landmark Points Detection Using Knowledge Distillation-Based Neural Networks}

\author{Ali \snm{Pourramezan Fard} (ali.pourramezanfard@du.edu)} 

\author{Mohammad H. \snm{Mahoor} (mmahoor@du.edu)}

\address{Department of Electrical and Computer Engineering, University of Denver, 2155 E Wesley Ave, Denver 80208, USA}

\begin{abstract}
Facial landmark detection is a vital step for numerous facial image analysis applications. Although some deep learning-based methods have achieved good performances in this task, they are often not suitable for running on mobile devices. Such methods rely on networks with many parameters, which makes the training and inference time-consuming. Training lightweight neural networks such as MobileNets are often challenging, and the models might have low accuracy. Inspired by knowledge distillation (KD), this paper presents a novel loss function to train a lightweight Student network (e.g., MobileNetV2) for facial landmark detection. We use two Teacher networks, a Tolerant-Teacher and a Tough-Teacher in conjunction with the Student network. The Tolerant-Teacher is trained using Soft-landmarks created by active shape models, while the Tough-Teacher is trained using the ground truth (aka Hard-landmarks) landmark points. To utilize the facial landmark points predicted by the Teacher networks, we define an Assistive Loss (ALoss) for each Teacher network. Moreover, we define a loss function called KD-Loss that utilizes the facial landmark points predicted by the two pre-trained Teacher networks (EfficientNet-b3) to guide the lightweight Student network towards predicting the Hard-landmarks. Our experimental results on three challenging facial datasets show that the proposed architecture will result in a better-trained Student network that can extract facial landmark points with high accuracy. 

\end{abstract}

\end{frontmatter}
% \clearpage

%\linenumbers

%% main text
\section{Introduction}\label{sec:intro}
% Face-alignment --->>>
Facial image alignment based on landmark points is a crucial step in many facial image analysis applications including face recognition~\cite{lu2015surpassing, soltanpour2017survey},
face verification~\cite{sun2014deep, sun2013hybrid}, face frontalization~\cite{hassner2015effective}, pose estimation~\cite{vicente2015driver}, and facial expression recognition~\cite{sun2014deep, zhao2003face}. The goal is to detect and localize the coordinates of predefined landmark points on human faces and use them for face alignment. In the past two decades, great progress has been made toward improving facial landmark detection algorithms' accuracy. However, most of the previous research does not focus on designing and/or training lightweight networks that can run on mobile devices with limited computational power.

While facial landmark points detection is still considered a challenging task for faces with large pose variations and occlusion~\cite{dong2018style, wu2018look}, recent methods have designed heavy models with a large number of parameters, which making them unsuitable for real-time applications. Moreover, with the growth of Internet-of-Things~(IoT), robotics, and mobile devices, it is vital to balance accuracy and model efficiency (i.e., computational time). Recently, deep learning-based methods have caught the attention of people in tackling this problem too. Among many lightweight neural network models, MobileNetV2~\cite{sandler2018MobileNetV2} is proven to be a good trade-off between accuracy and speed. However, because of the small number of network parameters, the face alignment task's accuracy using MobileNetV2~\cite{sandler2018MobileNetV2} might not be enough, especially when applied to faces with extreme poses or occlusions. 

Tan and Le~\cite{tan2019efficientnet} have recently proposed EfficientNet, a family of eight different networks designed to put a trade-off between the accuracy and model size. The designer of EfficientNet found a strong connection between the accuracy of a network and its depth, width, and resolution. Consequently, the proposed EfficientNet family is designed to be \textit{efficient}. In other words, EfficientNet family are designed to achieve good accuracy while they are relatively small means having a fewer number of network parameters and fast means having a smaller number of floating points operation (FLOPs).

% KD
Recently, knowledge distillation (KD) was utilized in image classification~\cite{hinton2015distilling,romero2014fitnets}, object detection~\cite{li2017mimicking}, and semantic segmentation~\cite{xie2018improving}. Initially, the idea was to train a lightweight network with acceptable accuracy by transferring features and knowledge generated by an ensemble network into the single smaller network \cite{modelcompression}. Later, Hinton et al.~\cite{hinton2015distilling} introduced the term \textit{knowledge distillation} as a technique to create a small model, called Student network, learned to generate the results that are created by a more cumbersome model, called Teacher network. 

% In other words, sometimes we face a problem in which a small model does not perform accurately enough. However, we can gain a more accurate performance using a cumbersome model (either a bigger model with more parameters or an ensemble of small models). In this situation, we can train the small model to mimic the cumbersome model results, called \textit{soft labels}, as predicting soft labels is easier than the original problem.

Inspired by the concept of KD, we propose a novel loss function called \textit{KD-Loss} to improve face alignment accuracy. Specifically, we propose a KD-based architecture using two different Teacher networks -- EfficientNet-B3~\cite{tan2019efficientnet} -- to guide the lightweight Student-Network, which is MobileNetV2~\cite{sandler2018MobileNetV2}, to better cope with the facial landmark detection task. Using the facial landmarks predicted by each of the Teacher networks, we introduce two ALoss functions. Being assisted by the ALoss, the KD-Loss considers the geometrical relation between the facial landmarks predicted by the Student and the two Teacher networks to improve the accuracy of the MobileNetV2~\cite{sandler2018MobileNetV2}. In other words, we proposed to use two independent sets of facial landmark points which are predicted by our Teacher network to guide the lightweight Student network towards better localization of the landmark points.

We train our method in two phases. In the first phase, we create Soft-landmarks inspired by ASM~\cite{cootes1998active}. Soft-landmarks are more similar to the Mean-landmark compared to the Hard-landmarks, which are the original facial landmarks. Hence, as a rule of thumb, it is easier for a lightweight model to predict the distribution of these Soft-landmarks compared to the original ground truth. We use this attribute to create a Teacher-Student architecture to improve the accuracy of the Student network. More clearly, in the first phase, we train one Teacher network using the Hard-landmarks and call it Tough-Teacher, and another Teacher network using the Soft-landmarks as the ground truth landmark points and call it Tolerant-Teacher.
Then, in the second phase, we use our proposed KD-Loss to transfuse the information gathered by both Teacher networks into the Student model during the training phase. Fig.~\ref{fig:general_framework} depicts a general architecture of our proposed training architecture. We tested our proposed method on the challenging 300W~\cite{sagonas2013300}, WFLW~\cite{wu2018look}, and COFW~\cite{burgos2013robust} datasets. The results of our experiments show that the accuracy of facial landmark points detection using MobileNetV2 trained using our KD-Loss approach is more accurate than the original MobileNetV2~\cite{sandler2018MobileNetV2}. The results are also comparable with state-of-the-art methods, while the network size is significantly smaller than most of the previously proposed networks.
\begin{figure}[t!]
  \centering
  \includegraphics[width=7.0cm]{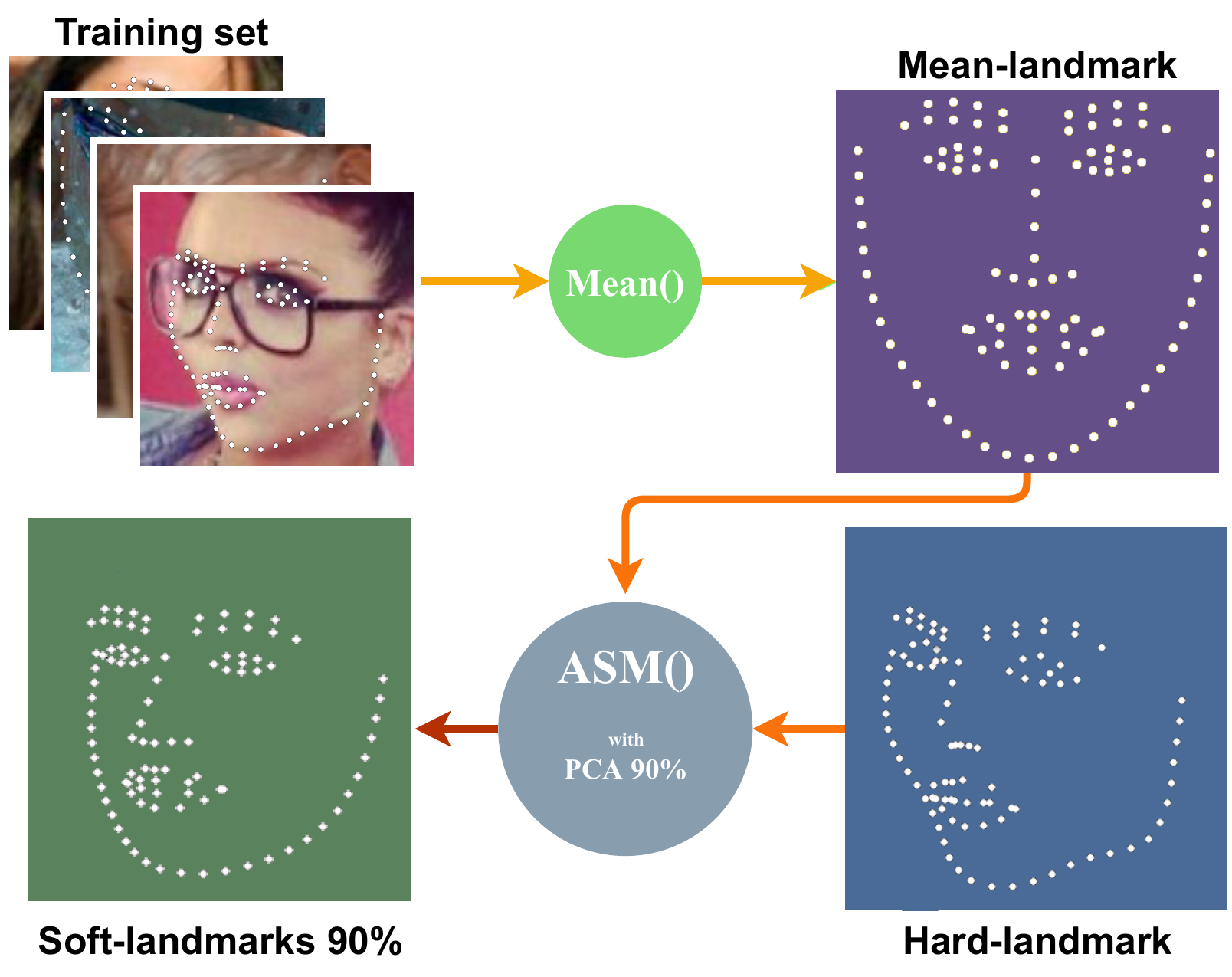}
  \caption{The process of creating Mean-landmark, and Soft-landmarks with different accuracy using Hard-landmarks.}
  \label{fig:soft_face_creationss}
\end{figure}

The contributions of our approach are summarized as follows. First, to the best of our knowledge, this is the first time the concept of KD is applied to a coordinate-based regression facial landmark detection model. Second, we proposed two different Teacher networks for guiding the Student network toward the ground truth landmark point. Third, different from the popular loss functions, the magnitude of our proposed ALoss can be either a \textit{positive} or a \textit{negative} number. Finally, using ALoss, we propose KD-Loss, which uses the geometrical relation between the Student and Teacher networks to improve accuracy in facial landmark detection. 

The remaining of this paper is organized as follows. Sec.~\ref{sec:lit} reviews the related work in facial landmark points detection. Sec.~\ref{sec:proposedModel} explains the details of the proposed method and the training process. Then, the evaluation of the method as well as the experimental results are provided in Sec.~\ref{sec:experiment}. Finally, Sec.~\ref{sec:conclusion} concludes with discussion on the proposed method and future research directions.  

% =======================RELATED WORKS====================================
\section{Related Work}
\label{sec:lit}
The facial landmark detection task dates back to over twenty years ago, when \textit{classical} methods (aka \textit{template-based methods}) were introduced. Active Shape Model (ASM)~\cite{cootes2000introduction} and Active Appearance Model (AAM)~\cite{cootes1998active, martins2013generative} are among the first methods for facial landmark detection. Based on these methods, Principal Components Analysis (PCA) is applied to simplify the problem and learn parametric features of faces to model facial landmarks variations. The model is iteratively fit into new instances. To match a 3D deformable face model to 2D images, \cite{martins2013generative} proposed a 2.5D AAM that combines a 3D metric Point Distribution Model (PDM) a 2D appearance model. The Constrained Local Model (CLM) proposed by Cristinacce and Cootes~\cite{cristinacce2006feature} and its various extensions including~\cite{asthana2013robust, baltruvsaitis20123d, saragih2011deformable, wang2008enforcing}, are among the most promising methods for face alignment. CLM models the face shapes with Procrustes analysis and principal component analysis. However, CLM methods are sensitive to occlusion as well as illumination when detecting landmarks in unseen datasets. To reduce the effect of outliers, Robust Cascade Pose Regression (RCPR)~\cite{burgos2013robust} was introduced to detect occlusions explicitly while using robust shape-indexed features. Another computationally lightweight method was Local Binary Features (LBF)~\cite{ren2014face}, which uses the locality principle to learn a set of highly discriminative local binary features from each facial landmarks independently. More over, Kazemi~\cite{kazemi2014one} proposed a gradient boosting framework using ensemble of regression trees for computationally-efficient face alignment. In their proposed framework, they first extract a sparse subset of intensity values form the input image, and then using cascade of regression trees to localize the facial landmark points.

\textit{Coordinate-based regression models} predict the facial landmark coordinates vector from the input image directly. Mnemonic Descent Method (MDM)~\cite{trigeorgis2016mnemonic} has utilized a recurrent convolutional network to detect facial landmarks. Feng et al.~\cite{feng2018wing} introduced Wingloss, a new loss function that is capable of overcoming the widely used L2 loss in conjunction with a strong data augmentation method as well as a pose-based data balancing (PDB). To ease the parts variations and regresses the coordinates of different parts, Two-Stage Re-initialization Deep Regression MODEL (TSR)~\cite{lv2017deep} splits face into several parts. \cite{zhang2018exemplar} proposed Exemplar-based Cascaded Stacked Auto-Encoder Network (ECSAN) for face alignment, which is utilized to handle partial occlusion in the image. To cope with self-occlusions and large face rotations, \cite{valle2019face} proposed a face alignment algorithm based on a coarse-to-fine cascade of ensembles of regression trees, which is initialized by robustly fitting a 3D face model to the probability maps produced by a pre-trained convolutional neural network (CNN). Guo~\cite{guo2019pfld} proposed a framework for practical face alignment, which estimates the rotation information during the train phase and use such information to better cope with the challenging faces with extreme pose, lighting and occlusion. Feng~\cite{feng2020rectified} proposed RWing loss, a piece-wise loss that amplifies the impact of the samples with small-medium errors, while rectifying the loss function for very small errors. More recently, Fard~\cite{fard2021asmnet} proposed a ASMNet, a lightweight multi-task network for jointly detecting facial landmark points as well as the estimation of face pose.

\begin{figure}[t!]
  \centering
  \includegraphics[width=\columnwidth]{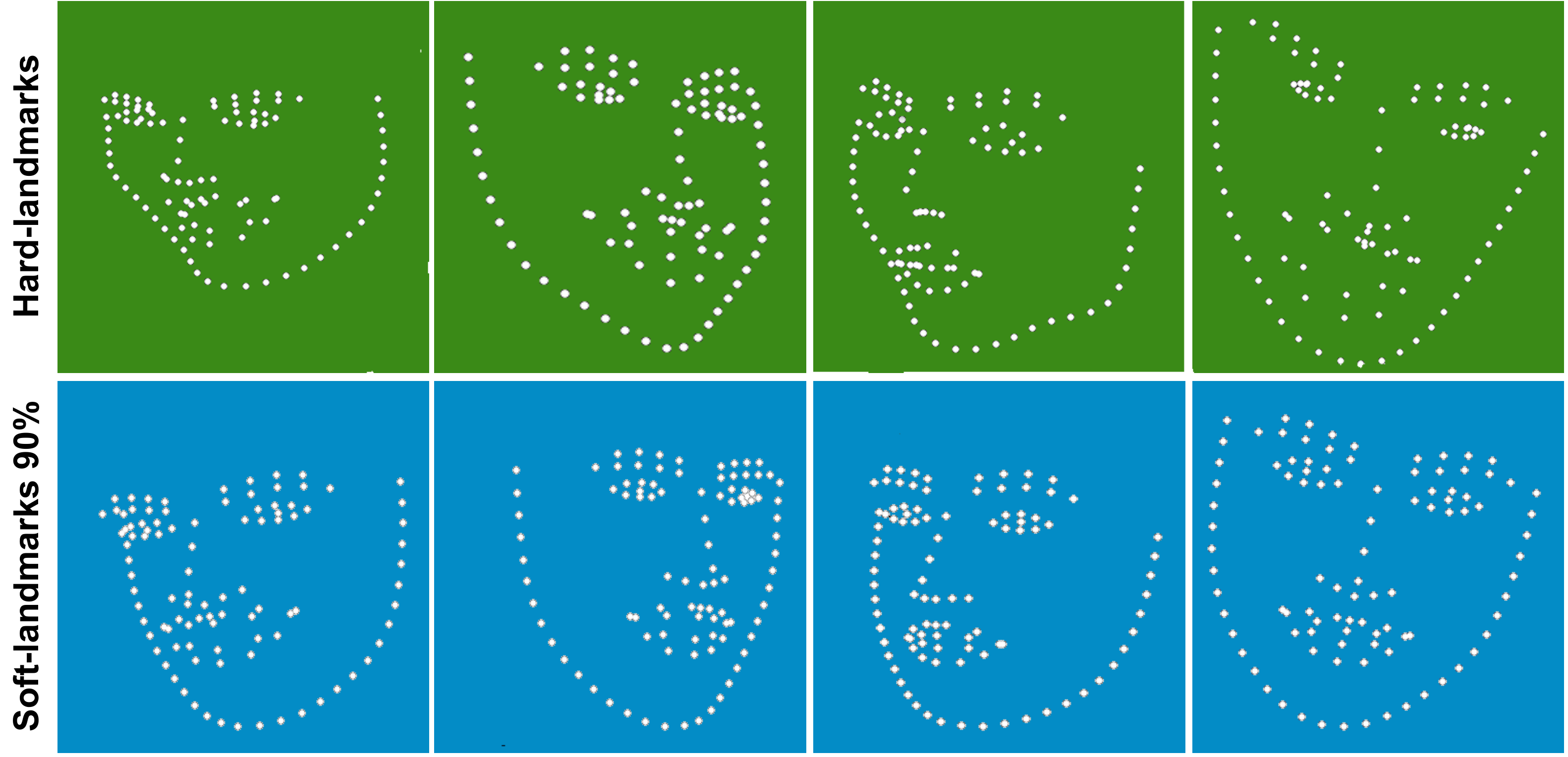}
  \caption{Examples of Soft-landmarks generated using $\tilde{m}=90\%$ of all the Eigenvectors, as well as Hard-landmarks.}
  \label{fig:soft_hard_samples}
\end{figure}
\begin{figure}[t]
  \centering
  \includegraphics[width=\columnwidth]{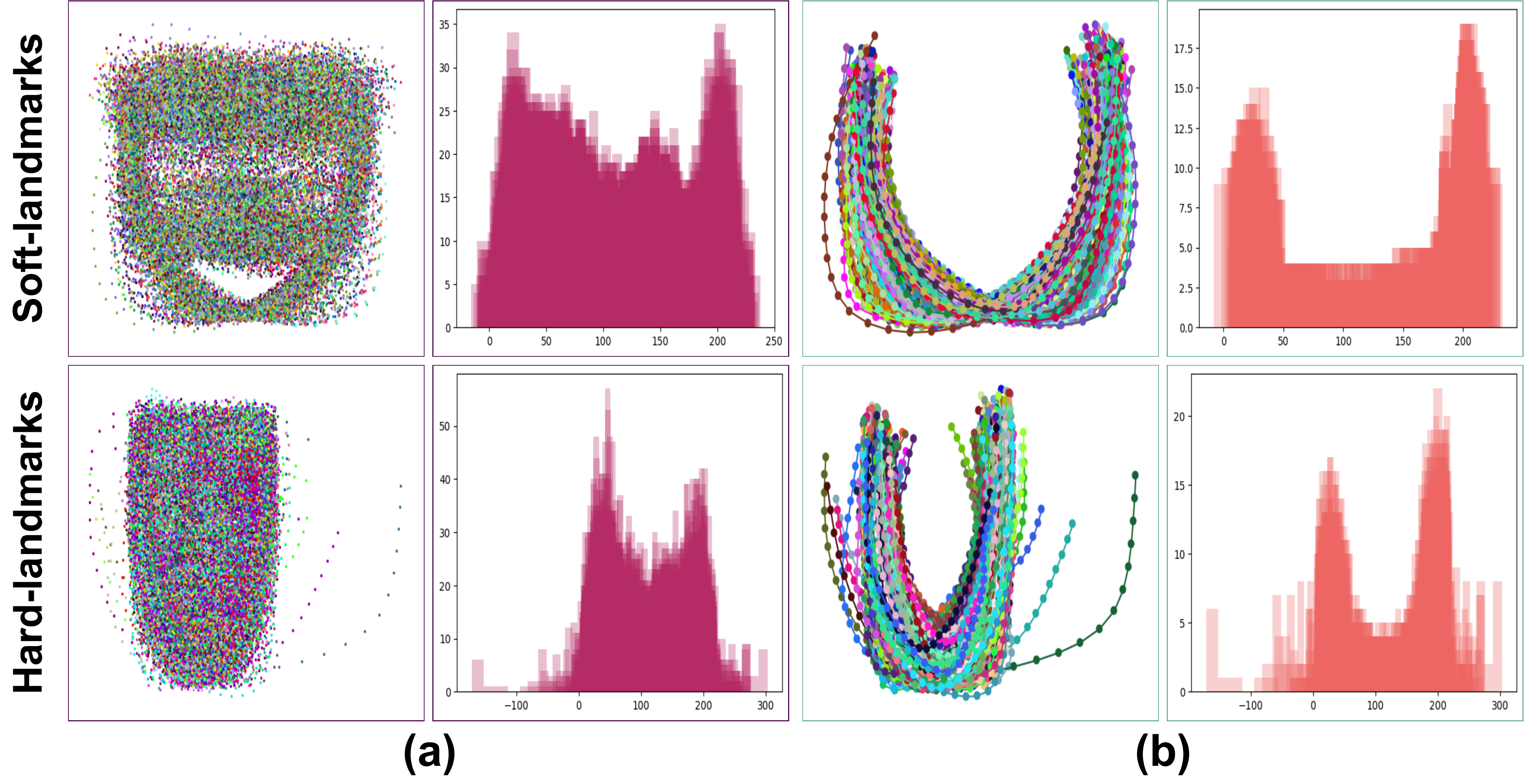}
  \caption{Distribution of Hard-landmarks and the Soft-landmarks created on WFLW~\cite{wu2018look} training set. (a) shows the distribution of all facial landmark, while for better visualization, (b) displays the landmark points belonging to the face boundary.}
  \label{fig:asm-Distribution}
\end{figure}

In \textit{heatmap-based regression models}, first, the likelihood heatmaps for each facial landmark are created, and then the network is trained to generate those heatmaps for each input image. A two-part network proposed by Yang~\cite{yang2017stacked}, including a supervised transformation to normalize faces and a stacked hourglass network~\cite{newell2016stacked}, is designed to predict heatmaps. In another work JMFA~\cite{deng2019joint} leveraged stacked hourglass network for multi-view face alignment, and it achieved state-of-the-art accuracy and demonstrated more accurate than the best three entries of the last Menpo Challenge~\cite{zafeiriou2017menpo}. LAB~\cite{wu2018look} proposed by Wu first expressed that the facial boundary line contains valuable information. Hence, they utilized boundary lines as the geometric structure of a face to help facial landmark detection. In another work, for a better initialization to Ensemble of Regression Trees (ERT) regressor, Valle~\cite{valle2018deeply} proposed a simple CNN to generate heatmaps of landmark locations. Additionally, \cite{sun2019high} introduced HRNet, a high-resolution network that is applicable in many Computer Vision tasks such as facial landmark detection and achieves a reasonable accuracy. \cite{iranmanesh2020robust} proposed an approach that provides a robust facial landmark detection algorithm that handles shape variations in facial landmark detection while the aggregating set of manipulated images to capture robust landmark representation. In another work, \cite{xiong2020gaussian} proposed the Gaussian heatmap vectors instead of the widely used heatmap for facial landmark points detection. More recently, to deal with the more challenging faces, \cite{mahpod2021facial} proposed a two-paired cascade subnetwork to generate heatmap and accordingly the coordinates of the facial landmark points.

Although heatmap regression models are more accurate than coordinate regression models, we follow the latter models since such models are significantly smaller in terms of both memory usage and the number of FLOPs, hence more suitable for mobile and embedded applications.

In addition, most of the previous work has proposed and/or utilized heavy networks with a large number of parameters and arithmetic operations. Consequently, such models are not applicable when utilized by embedded and mobile devices. In contrast, we propose a KD-based architecture and our novel KD-Loss to train lightweight models (e.g., MobileNetV2 having about 2.2 million parameters) that have significantly fewer parameters and arithmetic operations, while its accuracy is comparable with previous work. 

% =======================  PROPOSED MODEL =======================

%
\begin{figure}[t!]
  \centering
  \includegraphics[width=\columnwidth]{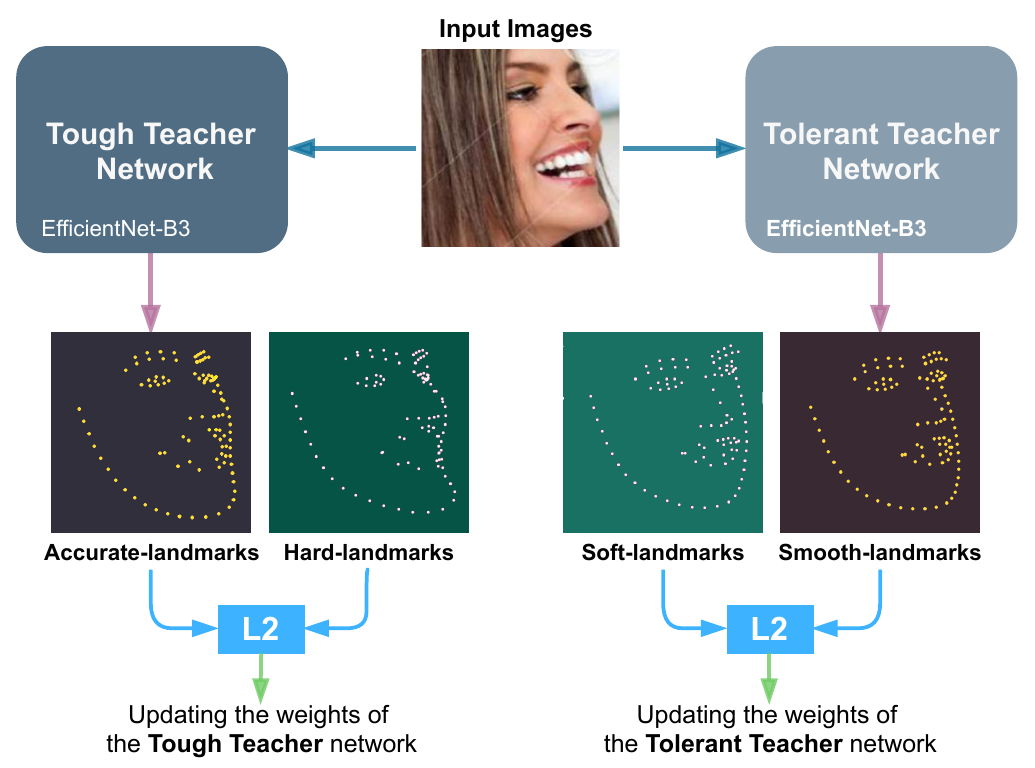}
  \caption{We train the Tough-Teacher, and the Tolerant-Teacher networks independently using the Hard-landmarks and the Soft-landmarks respectively utilizing the L2 loss. }
  \label{fig:general_framework}
\end{figure}

\section{Proposed Model}
\label{sec:proposedModel}
In this section, we first explain the process of creating the Soft-landmarks inspired from ASM. The Soft-landmarks are utilized for training the Tolerant-Teacher network. Then, we illustrate our proposed Student-Teacher architecture. After that, we explain our proposed KD-Loss function using the the proposed assistive loss functions.

\subsection{Soft and Hard Landmarks} \label{sec:lblCreation}
Inspired by ASM, we model a face shape object $\mathbf{f}$, which is a vector containing the coordinates of landmark points for each face, using Eq.~\ref{eq: ASM_1}:
\begin{equation} \label{eq: ASM_1}
\mathbf{f}_{(k\times1)} \approx \mathbf{\overline{f}}_{k\times1}  + \mathbf{V}_{k\times m}  \mathbf{b}_{m\times 1} 
\end{equation}
where $k$ is the number of the landmark points, $\mathbf{\overline{f}}$, the Mean-landmark, is the point-wise mean of all facial landmarks in the training set, and $\mathbf{V} = \{v_{1}, v_{2}, ... , v_{m}\}$ is a set containing \textit{m} Eigenvectors of the covariance matrix of all facial landmarks. $\mathbf{b}$ is also a $m-$dimensional vector given by Eq.~\ref{eq: ASM_2}:
\begin{equation} \label{eq: ASM_2}
\mathbf{b}_{m\times 1} = \mathbf{V}_{m\times k}^\intercal [ \mathbf{f}_{k\times 1} - \mathbf{\overline{f}}_{k\times 1} ]
\end{equation}
To ensure that the generated face is similar to the original face, $\mathbf{\tilde{b}}$ is defined by placing a restriction over $\mathbf{b}$ vector and limiting each of its elements to be between $-3\sqrt{\lambda_i}$ and $+3\sqrt{\lambda_i}$~\cite{cootes2000introduction}, where $\lambda_i$ is the statistical variance of the $i^{th}$ parameter of $b$. In Eq.~\ref{eq:ASM_3}, the new face shape $\mathbf{f\text{-}new}$ is created after applying this constraint:
\begin{equation} \label{eq:ASM_3}
\mathbf{f\text{-}new}_{k\times 1} = \mathbf{\overline{f}}_{k\times 1}  + \mathbf{V}_{k\times m} \mathbf{\tilde{b}}_{m\times 1} 
\end{equation}
Then we define the parameter $\tilde{m}$ as the proportion of the total number of Eigenvectors we use to generate the Soft-landmarks. In other words, $\tilde{m}$ define the similarity between the generated Soft-landmarks and the Hard-landmarks. Consequently, the smaller the parameter $\tilde{m}$, the fewer Eigenvectors we use to create $\mathbf{f\text{-}new}$ and thus, the generated $\mathbf{f\text{-}new}$ becomes more similar to the Mean-landmark. This effect is shown in Fig.~\ref{fig:soft_hard_samples} by providing some examples of Soft-landmarks.

Moreover, Fig.~\ref{fig:asm-Distribution} shows the distribution of the Hard-landmarks alongside Soft-landmarks created on WFLW~\cite{wu2018look} data set. While Fig.~\ref{fig:asm-Distribution}-(a) shows all the facial landmarks, in order to display the variations more clearly, in Fig.~\ref{fig:asm-Distribution}-(b) we only visualize the landmark points belonging to the face boundary. As Fig.~\ref{fig:asm-Distribution} shows, there are less variations in Soft-landmarks, and therefore it is easier for a deep neural network to learn such distributions. In this paper, we choose the parameter $\tilde{m}$ as 90\% of all the Eigenvectors and accordingly generate the Soft-landmarks using Eqs.~\ref{eq: ASM_1},~\ref{eq: ASM_2},~\ref{eq:ASM_3} (examples are shown in Fig.~\ref{fig:soft_face_creationss}).

\begin{figure}[t!]
  \centering
  \includegraphics[width=\columnwidth]{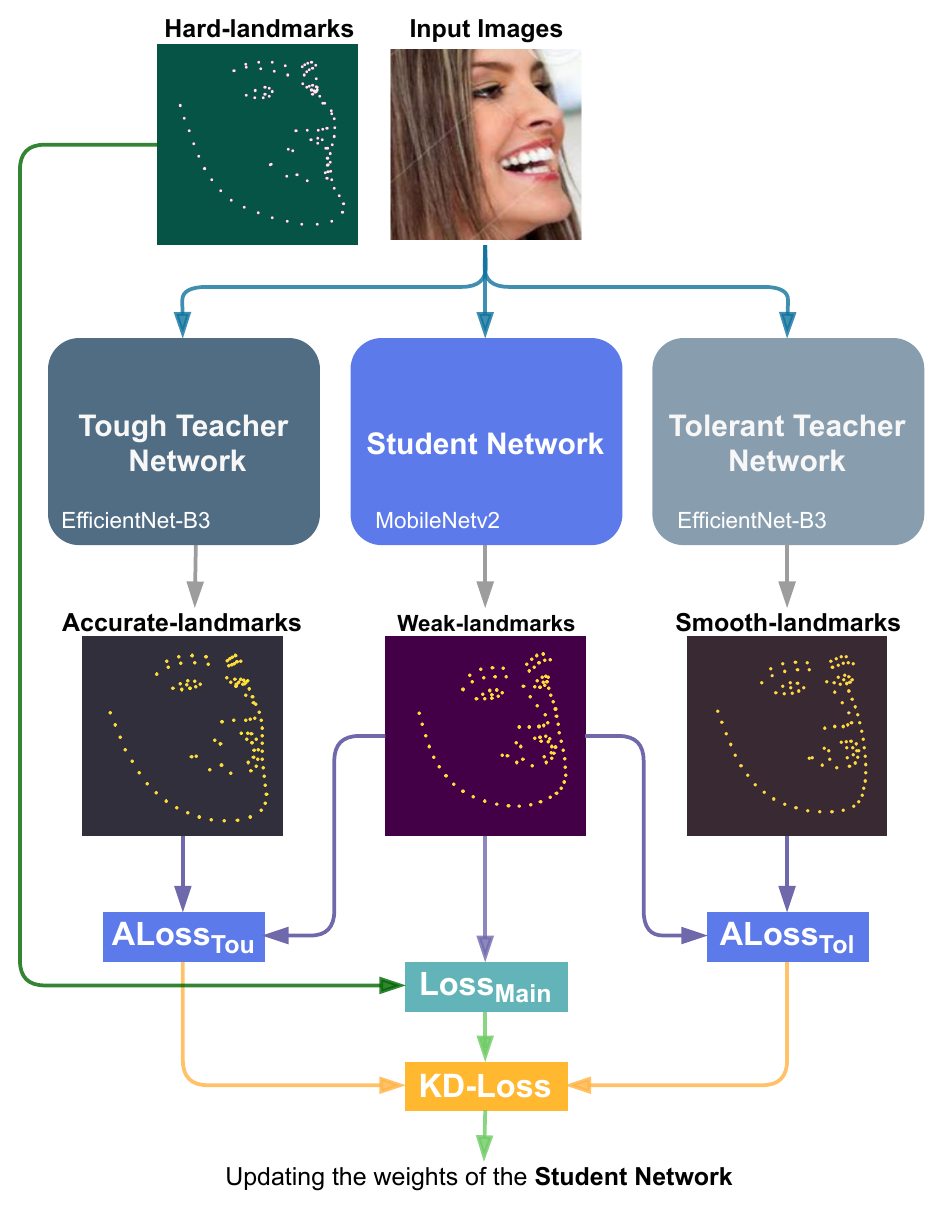}
  \caption{Proposed KD-based architecture for training the Student network. KD-Loss uses the knowledge of the previously trained Teacher networks by utilizing the assistive loss functions ALoss$_{Tou}$ and ALoss$_{Tol}$, to improve the performance the face alignment task.}
  \label{fig:general_framework}
\end{figure}

\subsection{Proposed Architecture} \label{sec:proposedArch}
Our proposed model consists of three main parts, two Teacher networks and a Student network. Being one of the best in the category of lightweight networks, we use MobileNetV2~\cite{sandler2018MobileNetV2} as the Student network. Furthermore, we choose EfficientNet-B3~\cite{tan2019efficientnet} as our Teacher network. In our proposed architecture, we have two Teacher networks: Tough-Teacher, which is trained using Hard-landmarks, and Tolerant-Teacher trained using Soft-landmarks. Since the variation of Soft-landmarks is smaller compared to the Hard-landmarks (see Sec.\ref{sec:lblCreation}), it is easier for deep neural networks to learn the face alignment task over the Soft-landmarks. However, the accuracy of the Soft-landmarks is lower in comparison to the original Hard-landmarks. We introduce the KD-Loss which uses the advantages of both Teachers to guide the Student network to learn the facial alignment task better. Tolerant-Teacher, trained on the Soft-landmarks has lower accuracy, but is easier to predict, while Tough-Teacher trained on the Hard-landmarks has higher accuracy, but is harder to predict.

Moreover, our proposed method consists of two phases. In the first phase, we independently train both Teacher networks using a standard L2 loss. In the second phase, we train the Student network using KD-Loss. More specifically, KD-Loss uses the landmark points generated by Tough and Tolerant Teachers to guide the Student towards learning the face alignment task more precisely. 

\subsection{Assistive Loss} \label{sec:assistedLoss}
We define ALoss to make the advantages of the geometrical knowledge of the Teacher networks. In other words, ALoss uses the facial landmark points predicted by each Teacher network to guide the Student network towards the ground truth. After training both the Tough-Teacher and Tolerant-Teacher on the training set independently, we have two different kinds of \textit{soft targets}: the Accurate-landmarks predicted by the Tough-Teacher as well as the Smooth-landmarks predicted by the Tolerant-Teacher. We define $P_{Gt}$ as an arbitrary facial landmark point from Hard-landmarks set, $P_{Ac}$, and $P_{Sm}$ the corresponding landmark points from the Accurate-landmarks and Smooth-landmarks sets respectively. Likewise, $P_{Pr}$ is the corresponding predicted points by the Student-Network. The idea behind the ALoss is to use the facial landmark points predicted using the Teacher networks as either a \textit{Positive} or \textit{negative} assistant. \textit{Positive} assistant means that the assistive loss function penalize the network to generate a landmark point which is close to the corresponding point predicted by the Teacher network, while the \textit{Negative} assistant means the assistive loss function penalize the Student network to predict a landmark point which is far from the corresponding landmark point predicted by the Teacher network.

\begin{figure}[t]
\centering
\includegraphics[width=\columnwidth]{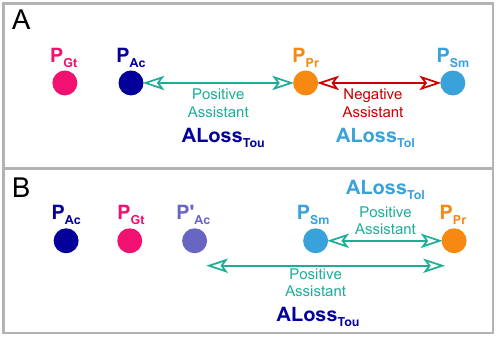}
\caption{Examples of using the assistive loss functions as a \textit{Positive} and \textit{Negative} assistant. A: ALoss$_{Tou}$ performs as a \textit{Positive} assistant, penalizing the Student network to predict $P_{Pr}$ to be close to $P_{Ac}$, while ALoss$_{Tol}$ is a \textit{Negative} assistant, penalizing the Student network to predict $P_{Pr}$ such that it is far from $P_{Sm}$. B: First, since $P_{Ac}$ is not between $P_{Pr}$ and $P_{Gt}$, we implicitly define $P'_{AC}$, and then consider both ALoss$_{Tou}$ and ALoss$_{Tol}$ as a \textit{Positive} assistant.}
\label{fig:l2_vs_l1}
\end{figure}

As an example, in Fig.~\ref{fig:l2_vs_l1}-A, in order to minimize the distance between $P_{Pr}$ and $P_{Gt}$, we use the ALoss$_{Tou}$ as a \textit{Positive} assistant, which means the ALoss$_{Tou}$ penalize the network to reduces the distance between $P_{Ac}$ and $P_{Pr}$. In contrary, the ALoss$_{Tol}$ is a \textit{Negative} assistant, which means the it penalize the network to predict $P_{Pr}$ to be far from $P_{Sm}$. More clearly, besides penalizing the Student network to learn the distribution of the Hard-landmarks, we guide it to learn both the distribution of the Accurate-landmarks as well as the Smooth-landmarks which are easier for a lightweight model. 

Moreover, to simplify and make the assistive loss function symmetric with respect to the coordinate of $P_{Gt}$, we need both $P_{Te}$ (a facial landmark point predicted using either of the Teacher networks) and $P_{Pr}$ to be in \textit{one} side of $P_{Gt}$. Therefore, we adapt the coordinate of $P_{Te}$ using Eq.~\ref{eq:adapt_teacher_coord}:
\begin{equation} \label{eq:adapt_teacher_coord}
    P_{Te} = P_{Gt} + sign(P_{Pr} - P_{Gt}) ~ |P_{Te} - P_{Gt}|
\end{equation}
As Fig.~\ref{fig:l2_vs_l1}-B shows, $P_{Ac}$ (or $P_{Sm}$) is not between $P_{Pr}$ and $P_{Gt}$, we use its symmetric point $P'_{Ac}$ calculated using Eq.\ref{eq:adapt_teacher_coord} to make the assistive loss a symmetric function with respect to $P_{Gt}$.
Then, we define parameter $\beta_{Te}$ with respect to the coordinate of both $P_{Gt}$ and $P_{Te}$ using Eq.~\ref{eq:Beta}:
\begin{equation} \label{eq:Beta}
    \beta_{Te} = P_{Gt}  +  \sigma_{Te} ~ sign(P_{Pr} - P_{Gt}) ~ |P_{Te} - P_{Gt}|
\end{equation}
$\beta_{Te}$ defines a threshold for the ALoss. As such, if the distance between $P_{Pr}$ and $P_{Gt}$ is smaller than the distance between $\beta_{Te}$ and $P_{Gt}$, then $P_{Pr}$ is considered as \textit{accurate enough}, and the ALoss gradually reduces the magnitude of the assistive loss. As shown in Eq.~\ref{eq:Beta}, we define $\beta_{Te}$ to be a portion ($\sigma_{Te}$) of the distance between $P_{Gt}$ and $P_{Te}$, where smaller values of $\sigma_{Te}$ penalize the Student network more. However, it forces the network to put too much effort toward improving the accuracy of landmark points which can reduce the accuracy in general. We choose $\sigma_{Te}$ as $0.4$, so the distance between $\beta_{Te}$ and $P_{Gt}$ is $40\%$ of the distance between $P_{Gt}$ and $P_{Te}$. Then the ALoss consider the prediction as \textit{accurate enough}.

In order to define the ALoss, we define an assistant weight function $\omega_{Te}$, which enables the ALoss to adjust its magnitude according to the coordinates of the $P_{Gt}$, $P_{Te}$ and $P_{Pr}$. We define $\omega_{Te}$ using Eq.~\ref{eq:weight_function}:
\begin{equation} \label{eq:weight_function}
\begin{adjustbox}{max width=230pt}$ 
    \omega_{Te}(p)= \left\{
    \begin{matrix*}[l]
    &1 &\forall~p \in ~\mathcal{R}_{P}           \\
    &-0.5 &\forall~p \in ~\mathcal{R}_{N}       \\
    &\frac{-0.5}{\beta_{Te} - P_{Gt}}       &\forall~p \in ~\mathcal{R}_{L}       \\
    \end{matrix*}\right.
$\end{adjustbox}
\end{equation}

We define $\mathcal{R}_{P}$ as a region in which ALoss acts as a \textit{Positive} assistant. We define this region using the relation between the location of the $P_{Gt}$, $P_{Pr}$, and $P_{Te}$ as follow in Eq.\ref{eq:rp}:
\begin{equation} \label{eq:rp}
\mathcal{R}_{P} : |P_{Pr} - P_{Gt}| - |P_{Te} - P_{Gt}| \geq 0
\end{equation}
which means for any predicted $P_{Pr}$, if the distance between $P_{Pr}$ and $P_{Gt}$ is greater than the distance between the $P_{Te}$ and $P_{Gt}$, ALoss acts as a \textit{Positive} assistant and penalize the Student network to predict $P_{Pr}$ to be close to $P_{Te}$. Similarly, we define $\mathcal{R}_{N}$ as a region in which ALoss acts as a \textit{Negative} assistant, and penalize the Student network to predict $P_{Pr}$ to be far from $P_{Te}$. We define this region as follow in Eq.\ref{eq:rn}:
\begin{equation} \label{eq:rn}
\mathcal{R}_{N} : 
\left\{
    \begin{matrix*}[l]
    & |P_{Te} - P_{Gt}| - |P_{Pr} - P_{Gt}|  \ge 0 \\
    & ~~~~~~~~~~~~~~~~~~\& \\
    & |P_{Pr} - P_{Gt}| - |B_{Te} - P_{Gt}| \ge 0 \\
    \end{matrix*}\right.
\end{equation}

Then, we define $\mathcal{R}_{L}$ as the \textit{Low Influence} region, meaning we consider the prediction of the landmark point $P_{Pr}$ as accurate enough if it is located in this region. Although ALoss acts as a \textit{Negative} assistant in this region, the magnitude of the ALoss decreases as $P_{Pr}$ get closer to $P_{Gt}$. we define $\mathcal{R}_{L}$ as follows in Eq.\ref{eq:rl}:
\begin{equation} \label{eq:rl}
\mathcal{R}_{L} : |B_{Te} - P_{Gt}| - |P_{Pr} - P_{Gt}| \geq 0
\end{equation}
In other words, we consider a predicted $P_{Pr}$ as accurate enough if it is located between the ground truth point, $P_{Gt}$, and $B_{Te}$ (see Eq.\ref{eq:Beta}). 
\begin{figure}[t]
\centering
\includegraphics[width=\columnwidth]{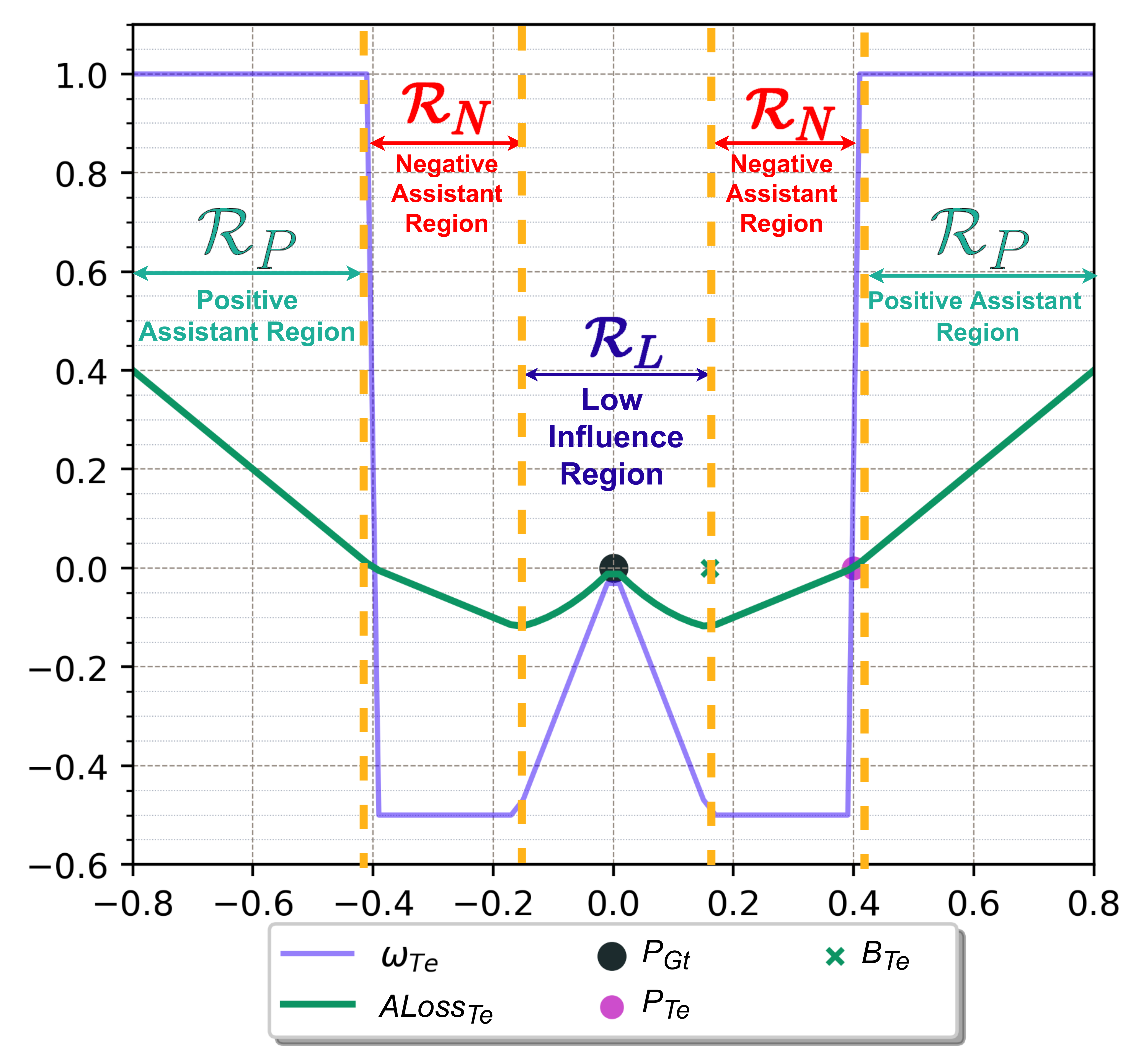}
\caption{ $\mathbf{\omega}_{Te}$ and ALoss for $P_{Gt}=0$, and $P_{Te}=0.4$.}
\label{fig:ALoss}
\end{figure}

Next, we define $ALoss_{Te}$ as the assistive loss with respect to $P_{Te}$ using Eq.~\ref{eq:ALoss}:
\begin{equation} \label{eq:ALoss}
    ALoss_{Te} = \omega_{Te} |P_{Te} - P_{Pr}| 
\end{equation}
We define Both $\omega_{Te}$ and consequently $ALoss_{Te}$ according to three different regions(see Fig.~\ref{fig:ALoss}, and Eqs.~\ref{eq:weight_function} and \ref{eq:ALoss}): \textbf{a)} the \textit{Positive Assistant} region, $\mathcal{R}_{P}$ where the value of $\omega_{Te}$ is $1$, a \textit{positive} number, and hence, $ALoss_{Te}$ is also a \textit{positive assistive} loss, meaning that the closer $P_{Pr}$ is to $P_{Te}$, the smaller the value of the $ALoss_{Te}$. \textbf{b)} the \textit{Negative Assistant} region, $\mathcal{R}_{N}$, where $P_{Pr} \in [\beta_{Te}, P_{Te})$. In this region, $\omega_{Te}$ is define as $-0.5$, which is a \textit{negative} number. Consequently, $ALoss_{Te}$ is defined as a \textit{negative assistive} loss, meaning that the \textit{further} the $P_{Pr}$ is from the $P_{Te}$, the smaller the value of $ALoss_{Te}$ becomes. In other words, in this region we design $ALoss_{Te}$ to train the Student network to use the coordinates of $P_{Te}$ and tries to increase the distance between $P_{Pr}$ and $P_{Te}$. \textbf{c)} the \textit{Low Influence} region, $\mathcal{R}_{L}$, where $P_{Pr} \in [P_{Gt}, \beta_{Te})$. As shown in Eq.~\ref{eq:weight_function}, in this region $\omega_{Te}$ is a linear function with negative slope where its minimum value, $-0.5$, is at $\beta_{Te}$ and its maximum value, $0$, is at $P_{Gt}$. If the predicted point $P_{Pr}$ is closer to $P_{Gt}$ than $\beta_{Te}$, we consider the prediction as \textit{good enough}, and we design $\omega_{Te}$ to gradually reduce the loss magnitude. Furthermore, $\omega_{Te}$ is designed such that $ALoss_{Te}$ be continuous and its value is $0$ when $P_{Pr}$ is equal to $P_{Gt}$. In practice, the facial landmark points used for training the models are zero-centered normalized, which means for any arbitrary facial landmark point $P_{x,y}$, both \textit{x} and \textit{y} coordinates are in $[-0.5, 0.5]$ range. 

As Eq.~\ref{eq:ALoss} shows, $ALoss_{Te}$ is a piece-wise continuous function. In addition, it is a linear function in $\mathcal{R}_{P}$, and $\mathcal{R}_{N}$ regions, and a quadratic function in $\mathcal{R}_{L}$ region. Fig.~\ref{fig:ALoss} shows $\omega_{Te}$ and $ALoss_{Te}$ functions, for $P_{Gt}=0$ and $P_{Te}=0.4$.

\begin{figure}[t]
\centering
\includegraphics[width=7.5cm]{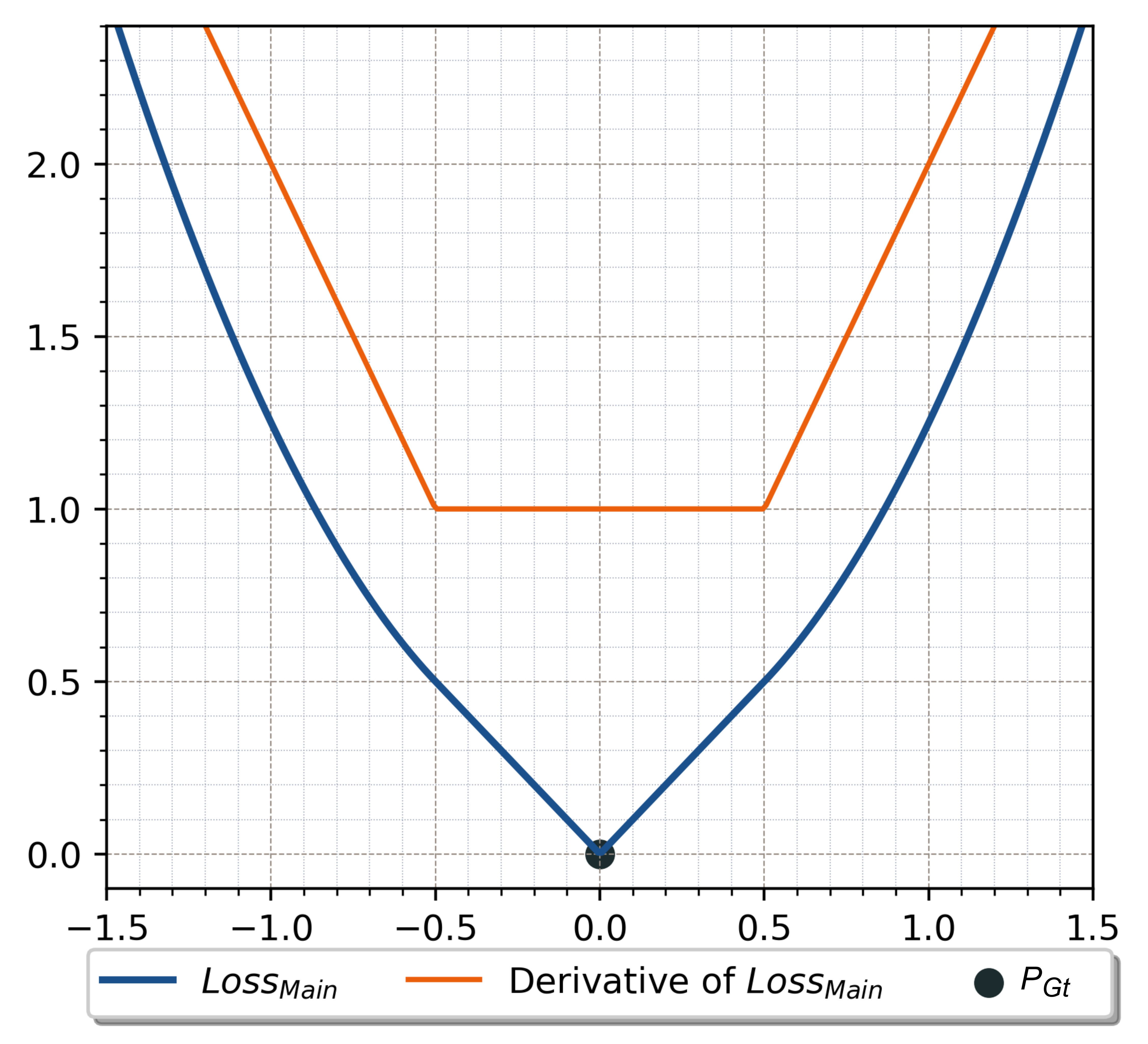}
\caption{ Loss\textsubscript{Main} for $P_{Gt}=0$.}
\label{fig:main_loss}
\end{figure}
\begin{figure*}[t!]
  \centering
  \includegraphics[width=14cm]{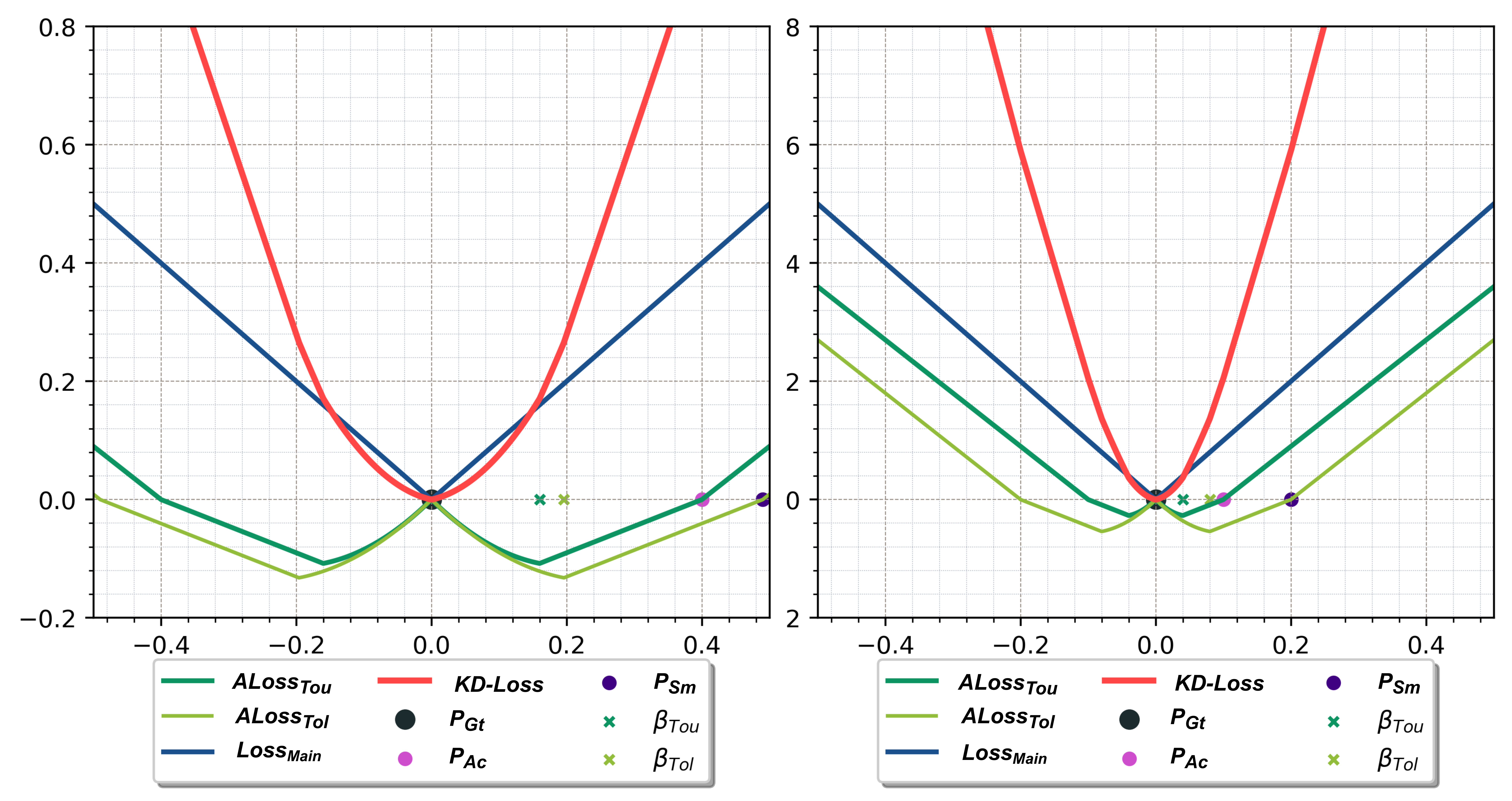}
  \caption{Two examples of Proposed KD-Loss. The larger the distance between $P_{Gt}$ and $P_{Ac}$ (or $P_{Sm}$), the greater the absolute value of the corresponding $ALoss_{Tou}$ (or $ALoss_{Tol}$) in $\mathcal{R}_{N}$. As a consequence, the effect of the negative assistive loss is more in the left figure, compared to the right figure.}
  \label{fig:kd_loss_and_weight}
\end{figure*}

\subsection{Proposed KD-Loss} \label{sec:knowledgeTransfer}
Hinton~\cite{hinton2015distilling} used the term \textit{Knowledge Distillation}, as they created \textit{soft targets} as the class probabilities generated by the Teacher model, which replaced the \textit{soft max} layer of the network with a \textit{logit} layer. In contrast, our novel KD-Loss and the proposed a Teacher-Student architecture are different than the standard KD concept in 2 ways: first, while KD is mostly used for classification tasks, to the best of our knowledge, this is the first time KD is used in a coordinate regression task. Second, in the original KD concept, the Student network tries to \textit{mimic} the output results of the Teacher network while in our proposed KD-Loss, we provide two different assistant loss functions, $ALoss_{Tou}$ and $ALoss_{Tol}$ corresponding to the facial landmark points predicted by the Tough and the Tolerant Teachers respectively. Accordingly, opposite to the original KD idea, where the Student network tries to mimic the results predicted by the Teacher network, our Student-Network, guided by KD-Loss, uses the coordinates of facial landmark points predicted by the Teacher networks to better predict the ground truth points. Third, we utilize two different Teacher networks, Tough-Teacher and call its predicted facial landmark points as Accurate-landmarks, as well as the Tolerant-Teacher and call its corresponding predicted facial landmark points as Smooth-landmarks. Comparing with the ground truth set, called as the Hard-landmarks, Accurate-landmarks are more accurate than the Smooth-landmarks, while having more complex distribution.

In order to define the KD-Loss, we first need to define the formal equation for the both assistive loss functions. We define $P_{GT}[i,n]$ as the ground truth coordinate of the $i^{th}$ landmark point of the $n^{th}$ image in the training set. Likewise, we define $P_{Pr}[i,n]$, $P_{Ac}[i,n]$, and $P_{Sm}[i,n]$ as the corresponding landmark points from the predicted, accurate, and smooth landmark points set respectively. In addition, for each facial landmark point $P_{GT}[i,n]$, we define $\omega_{Tou}[i,n]$ and $\omega_{Tol}[i,n]$ as the assistant weight functions corresponding to the Tough, and Tolerant Teachers, respectively (see Eq.~\ref{eq:weight_function}).

Then, we define $ALoss_{Tou}[i,n]$ and $ALoss_{Tol}[i,n]$, the assistive loss functions corresponding to the Tough, and Tolerant Teachers, respectively using Eq.~\ref{eq:ALoss}. Accordingly, we define the assistant loss functions for all facial landmark points in the training set using Eq.~\ref{eq:Total_MDAL}:
\begin{equation} \label{eq:Total_MDAL}
\begin{split}
    ALoss_{Tou} = \frac{1}{N~k} \sum_{n=1}^{N} \sum_{i=1}^{k} ALoss_{Tou}[i,n] \\
    ALoss_{Tol}=  \frac{1}{N~k} \sum_{n=1}^{N} \sum_{i=1}^{k} ALoss_{Tol}[i,n]
\end{split}
\end{equation}
where $k$ is the number of the facial landmark points in a face, and \textit{N} is the number of all samples in the training set. Moreover, we define $Loss_{Main}$ in Equations.~\ref{eq:delta}, ~\ref{eq:mainLoss}:
\begin{equation} \label{eq:delta}
\Delta[i,n] = |P_{Gt}[i,n] - P_{Pr}[i,n]|
\end{equation}

\begin{equation} \label{eq:mainLoss}
Loss_{Main} = 
    \left\{
    \begin{matrix*}[l]
     & \frac{1}{N~k} \sum_{n=1}^{N} \sum_{i=1}^{k} \Delta[i,n] 
     &\text{If: ~} \Delta[i,n] \leq 0.5 \\ 
     & \frac{1}{N~k} \sum_{n=1}^{N} \sum_{i=1}^{k} \Delta^2[i,n] + C
     &\text{otherwise}
    \end{matrix*}\right.
\end{equation}
where $C=0.25$ is defined to smoothly connect the two pieces together. As shown in Fig.~\ref{fig:main_loss}, for any facial landmark point $P_{Gt}[i,n]$, we define $Loss_{Main}$ as a continuous piece-wise function with respect to the parameter $\Delta$. Accordingly, for any facial landmark points if $\Delta[i,n]$ is greater than $0.5$, we define $Loss_{Main}$ as L2 loss. On the contrary, for $\Delta[i,n]$ smaller than $0.5$, we define $Loss_{Main}$ as L1 loss. To make the most advantages of both L2 and L1 loss, we define our proposed $Loss_{Main}$ as a combination of both loss functions.

L2 loss ($y = x^2$) penalize the model more for the large. Since its derivative ($y' = 2x$) is a linear function of the errors, the larger the error, the larger the magnitude of the derivative (the influence of the loss function). For small errors (errors that are smaller than 1), its influence becomes very low, leading the network to focus more on large errors, while neglecting small errors.
On the contrary, the influence of L1 loss ($y= x$) is 1 ($y = 1$), meaning L1 is not sensitive to the large errors. Accordingly, the magnitude and the influence of the errors in L1 loss is larger compared to L2 loss for small errors. Given this, we define $Loss_{Main}$ with the parameter $\Delta=0.5$ (in Sec.\ref{sec:assistedLoss}, we discussed that coordinates of all facial landmarks are in [-0.5, 0.5]) to both be sensitive to small and large errors.

Then, we define KD-Loss as a linear combination of $ALoss_{Tou}$, $ALoss_{Tou}$ and $Loss_{Main}$ in Eq.~\ref{eq:KD_based_loss}, and also depict it in Fig.~\ref{fig:kd_loss_and_weight}:
\begin{equation} \label{eq:KD_based_loss}
\begin{split}
    \text{KD-Loss} = \phi \times Loss_{Main} + ALoss_{Tou} + ALoss_{Tou}
\end{split}
\end{equation}
As we have two assistant loss functions, we the parameter $\phi$ to be \textit{2} to equalize the effect of $Loss_{Main}$ with the sum of $ALoss_{Tou}$ and $ALoss_{Tou}$. Moreover, since we have trained both Tough and Tolerant Teachers separately before, the coordinates of the points predicted by them is in $[-0.5, +05]$. Besides, According to Equations.~\ref{eq:weight_function} and~\ref{eq:ALoss}, the minimum value for ALoss will be at its corresponding $\beta$. By defining the parameter $\sigma=0.4$ (see Eq.~\ref{eq:Beta}), we ensure that the proposed KD-Loss will never be negative in its domain of declaration.

According to Eq.~\ref{eq:ALoss} (also is shown in Fig~\ref{fig:kd_loss_and_weight}), the further the points predicted by the Teacher from its corresponding ground truth point, the higher the \textit{negative} value of the corresponding assistant loss. In other words, the distance between the points predicted by Teacher networks from their corresponding ground truth can imply how hard is the distribution of that point. Accordingly, when the distribution of the ground truth points is hard, KD-Loss tries to guide the network by increasing the effect of the assistant loss functions in order to put more attention on corresponding \textit{Smooth} or \textit{Accurate} faces rather than Hard-landmarks.

While for the L2 loss, the curvature of the loss function corresponds to the difference between the ground truth points and the corresponding predicted points, the curvature of KD-Loss is defined according to $\beta_{Tou}$ and $\beta_{Tol}$ (see Eq.~\ref{eq:Beta}). For each point $P_{Gt}[i,n]$, we define the parameters $\beta_{Tou}[i,n]$ and $\beta_{Tol}[i,n]$ with respect to $P_{Ac}[i,n]$ and $P_{Sm}[i,n]$ respectively. The relation between the $\beta_{Tou}[i,n]$ and $\beta_{Tol}[i,n]$, and the corresponding points predicted by Teacher networks, $P_{Ac}[i,n]$ and $P_{Sm}[i,n]$, adjusts the curvature of the corresponding $KD-Loss[i,n]$. This means that, the further the distance between the points predicted by Teacher networks from their corresponding ground truth, the further the corresponding $\beta^n\_Tou_{i}$ and $\beta^n\_Tol_{i}$, and thus the wider the \textit{Low Influence} region (see Sec.~\ref{sec:assistedLoss}), and the smoother the curvature of $KD-Loss[i, n]$. It is also shown in Fig.~\ref{fig:kd_loss_and_weight}, that the curvature of the KD-Loss relates to the both $\beta_{Tou}$ and $\beta_{Tol}$.

%  =======================  Experimental Results =======================
% 
% 
\section{Experimental Results}
\label{sec:experiment}
In this section, we first explain the training phase and the datasets that we used in evaluating our proposed model. We then describe the test phase as well as the implementation details and the evaluation metrics. Lastly, we present the results of facial landmark points detection using our proposed KD-Loss method.

\subsection{Datasets} \label{sec:datasets} 
We conducted the training and evaluation of our models on three popular and challenging datasets: 300W~\cite{sagonas2013300}, COFW~\cite{burgos2013robust}, and WFLW~\cite{wu2018look}. 

\textbf{300W:} Following the protocol described in~\cite{ren2014face}, we train our networks using all 3148 68-points manually annotated faces. We also perform the testing on the common subset (554 images), the challenging subset (135 images), and the full set, which is the sum of the challenging subset and the common subset with 689 images consisting of 2000 images from the training subset of HELEN~\cite{le2012interactive} dataset, 337 images from the full set of AFW~\cite{zhu2012face} dataset, and 811 images from the training subset of LFPW~\cite{belhumeur2013localizing} dataset with a 68-point annotation. Images in HELEN, LFPW, and AFW datasets are collected in the wild environment; as a result, expression and large pose variations, as well as partial occlusions, may exist. For the test phase, the images are divided into two subsets, such that the common subset contains 554 images from LFPW~\cite{belhumeur2013localizing} and HELEN~\cite{le2012interactive}, and the challenging subset contains 135 images from IBUG~\cite{sagonas2013300}. The sum of the challenging subset and the common subset is considered as the full subset, containing 689 images.

\textbf{COFW:} This dataset has 1345 facial images as training and 507 facial images as the testing set. The dataset provides us with facial images having large pose variations plus heavy occlusions. Each image in the COFW~\cite{burgos2013robust} dataset has 29 manually annotated landmarks.

\begin{table}[t!]
\caption{NME (in \%) and failure rate of 29-point landmarks detection on COFW~\cite{burgos2013robust} dataset.}
\label{tbl:tbl_results_cofw}
\centering
\resizebox{\columnwidth}{!}{
\begin{tabular}{l c c }
\hline
Method                                   & NME                      & FR  \\ \hline \hline 

SFPD~\cite{wu2017simultaneous}            & 6.40                  & -         \\ 
DAC-CSR~\cite{feng2017dynamic}            & 6.03                  & 4.73      \\ 
CNN6 (Wing + PDB)~\cite{feng2018wing}     & 5.44                  & 3.75      \\ 
ResNet50 (Wing + PDB)~\cite{feng2018wing} & 5.07                  & 3.16      \\ 
LAB~\cite{wu2018look}                     & 3.92                  & 0.39      \\ 
ODN~\cite{zhu2019robust}                  & 5.30                  &  -        \\ 
HRNetV2~\cite{sun2019high}                & 3.45                  & 0.19      \\
ResNet50-FFLD~\cite{yan2020fine}          & 5.32                  & -       \\ 
GV(HRNet)~\cite{xiong2020gaussian}        & 3.37                  & 0.39    \\ 
\multicolumn{3}{l}{}
\\[-1em] \hline

mnv2                                    & 5.04                    & 3.74    \\
\textbf{mnv2\textsubscript{KD}  }       & 4.11                    & 2.36    \\
efn                                     & 3.81                    & 1.97    \\

\hline
\end{tabular}}
\end{table}

\textbf{WFLW:} This is another widely used facial dataset, which contains 7500 images for training, and 2500 images for testing and recently has been proposed based on WIDER FACE~\cite{yang2016wider}. Each image in this dataset contains 98 manually annotated landmarks. This dataset consists of 6 subsets, including 326 large pose images, 314 expression images, 698 illumination images, 206 make-up images, 736 occlusion images, and 773 blurred images. Consequently, it is possible to validate the robustness of the proposed model against each different condition.

\subsection{Evaluation Metrics} \label{sec:vel_metrics}
We evaluate our proposed KD-based architecture using normalized mean error~(NME), failure rate (FR), and the area-under-the-curve~(AUC)~\cite{yang2015empirical}. For the NME, we use “inter-ocular” distance (the distance between the outer-eye-corners) as the normalizing factor followed by MDM~\cite{trigeorgis2016mnemonic} and\cite{sagonas2013300}. On COFW~\cite{burgos2013robust}, and WFLW~\cite{wu2018look}, we calculate the FR, which is the proportion of failed detected faces for a maximum error of 0.1. We report the AUC for WFLW~\cite{wu2018look} as well.
\begin{table}[t!] 
\caption{NME (in \%) of 68-point landmarks detection on 300W~\cite{sagonas2013300}.}
\label{tbl:tbl_results_300w}
\centering
\small
\resizebox{\columnwidth}{!}{\begin{tabular}{l c c c}
\hline
\multirow{2}{*}{Method} & \multicolumn{3}{c}{Normalized Mean Error}       \\ \cline{2-4} 
                            & Common & Challenging  & Fullset \\ \hline\hline
DAN~\cite{kowalski2017deep}   & 3.19     & 5.24        & 3.59 \\ 
DSRN~\cite{miao2018direct}    & 4.12     & 9.68        & 5.21 \\ 
RCN~\cite{honari2016recombinator}  & 4.67  & 8.44    & 5.41     \\ 
CPM~\cite{dong2018supervision}    & 3.39           & 8.14        & 4.36 \\ 
PCD-CNN~\cite{kumar2018disentangling}    & 3.67   & 7.62    & 4.44   \\
ODN~\cite{zhu2019robust}    & 3.56   & 6.67    & 4.17     \\ 
SAN~\cite{dong2018style}    & 3.34    & 6.60        & 3.98     \\ 
LAB~\cite{wu2018look}    & 2.98           & 5.19        & 3.49  \\ 
DCFE~\cite{valle2018deeply}           & 2.76            & 5.22        & 3.24  \\ 
PFLD 1X~\cite{guo2019pfld}                  & 3.01          & 5.08           & 3.40    \\ 
LRefNets~\cite{su2019efficient}           & 2.71            & 4.78        & 3.12    \\
HRNetV~\cite{sun2019high}     & 2.87            & 5.15        & 3.32      \\
AWing~\cite{wang2019adaptive}    & 2.72            & 4.52        & 3.07    \\
3DDE~\cite{valle2019face}          & 2.69            & 4.92        & 3.13 \\  
ResNet50-FFLD~\cite{yan2020fine}  & 3.06             & 5.44       &3.50   \\  GEAN~\cite{iranmanesh2020robust}   & 2.68            & 4.71        & 3.05    \\ 
GV(HRNet)~\cite{xiong2020gaussian}  & 2.62            & 4.51        & 2.99 \\
GV(HRNet)~\cite{xiong2020gaussian}  & 2.62            & 4.51        & 2.99 \\
CCNN ~\cite{mahpod2021facial}                    & 3.23            & 3.99        & 3.44             \\ \hline

mnv2                                & 3.93     & 6.84       & 4.50          \\
\textbf{mnv2\textsubscript{KD} }    & 3.56     & 6.13       & 4.06       \\ 
efn                                 & 3.34     & 5.80       & 3.82         \\\hline

\end{tabular}}
\end{table}

\subsection{Implementation Details} \label{sec:imp_detail}
For each training image, we first crop and extract the faces. For 300W~\cite{sagonas2013300}, no bounding boxes are provided, and for COFW~\cite{burgos2013robust} and WFLW~\cite{wu2018look}, we figure out that the provided bounding boxes are not accurate enough. Accordingly, we generate the new bounding boxes based on the ground truth facial landmark points for each image. We then expand the bounding boxes randomly up 10\%. The next step is to resize each facial image to $224\times224$ pixels. To improve our models' accuracy and robustness, we augment each facial image multiple times in terms of brightness, contrast, and color modification as well as adding Gaussian noises. Moreover, we randomly rotate each image by ±45 degree and flip horizontally with the probability of 50\%. In the first stage, we train the both Tough and Tolerant Teacher networks independently for about 250 epochs with a batch size of 40 using L2 loss. Then, we train the Student network for about 250 epochs with a batch size of 70, using the proposed KD-Loss. We use Adam optimizer~\cite{kingma2014adam} with a learning rate $10^{-3}$, $\beta_1 = 0.9$, $\beta_2 = 0.999$, and $decay = 10^{-6}$. We use TensorFlow library to implement our codes and run them on a NVidia 1080Ti GPU. The code is publicly available online on \href{https://github.com/aliprf/KD-Loss}{GitHub}.

\begin{table}[t!] 
\caption{Comparison of the NME (in \%) of lightweight models in landmarks localization on 300W~\cite{sagonas2013300} dataset.}
\label{tbl:tbl_results_300w_small}
\centering
\small
\resizebox{\columnwidth}{!}{
{\begin{tabular}{l c c c }
\hline 
\multirow{2}{*}{Method} & \multicolumn{3}{c}{NME}       \\ \cline{2-4} 
                                            & Common & Challenging  & Fullset \\ \hline\hline
                     & \multicolumn{3}{c}{inter-ocular normalization}  \\
                                            
res\_loss~\cite{ning2020cpu}                 & -            & -             & 4.93 \\ 
ASMNet~\cite{fard2021asmnet}                & 4.82          & 8.2           & 5.50    \\ 
MobileNet+ASMLoss~\cite{fard2021asmnet}     & 3.88          & 7.35          & 4.59    \\ 
% PFLD 0.25X~\cite{guo2019pfld}               & 3.03          & 5.15           & 3.45    \\ 

\textbf{mnv2\textsubscript{KD} }            & 3.56          & 6.13          & 4.06       \\ 

\hline 
                    & \multicolumn{3}{c}{inter-pupil normalization}          \\

LBF~\cite{ren2014face}                   & 4.95    & 11.98   & 6.32         \\
LBF fast~\cite{ren2014face}              & 5.38    & 15.50   & 7.37         \\
CFSS~\cite{zhu2015face}                 &4.73       &9.98   &5.76           \\
3DDFA~\cite{zhu2016face}                &6.15       &10.59   &7.01          \\
DOF~\cite{wu2021design}                 & 4.86     & 9.13   & 5.55          \\
MuSiCa68~\cite{shapira2021knowing}      & 4.63     & 8.16   & 5.32          \\
G\&LSR$\omega$ ~\cite{shao2021robust}   & 4.52     & 7.82   & 5.17          \\

\textbf{mnv2\textsubscript{KD}}          & 4.97     & 8.90   & 5.66

\\ \hline
\end{tabular}}}
\end{table}

\begin{table*}[t]
\caption{NME (in \%), FR (in \%), and AUC of 98-point landmarks detection on WFLW~\cite{wu2018look} dataset.}
\label{tbl:tbl_results_wflw}
\centering
\small
\resizebox{15.0cm}{!}{

\begin{tabular}{ c l c c c c c c c}
\hline
Metric & Method & Test set & Pose  & Expression & Illumination &Make-Up & Occlusion & Blur                                                                              \\ \hline \hline
\multirow{2}{*}{NME} & \begin{tabular}[c]{@{}l@{}}
DVLN~\cite{wu2017leveraging}\\ LAB~\cite{wu2018look}\\ ResNet50(Wing+PDB)~\cite{feng2018wing} \\3DDE~\cite{valle2019face}   \\GV(HRNet)~\cite{xiong2020gaussian}
\end{tabular} & 
\begin{tabular}[c]{@{}l@{}} 6.08\\   5.27\\ 5.11 \\4.68 \\4.33 \end{tabular}  & 
\begin{tabular}[c]{@{}l@{}} 11.54\\ 10.24\\ 8.75 \\8.62 \\7.41 \end{tabular}&
\begin{tabular}[c]{@{}l@{}} 6.78\\  5.51\\  5.36 \\5.21 \\4.51 \end{tabular}&
\begin{tabular}[c]{@{}l@{}} 5.73\\  5.23\\  4.93 \\4.65 \\4.24 \end{tabular} &
\begin{tabular}[c]{@{}l@{}} 5.98\\  5.15 \\ 5.41 \\4.60 \\4.18 \end{tabular}&
\begin{tabular}[c]{@{}l@{}} 7.33\\  6.79\\  6.37 \\5.77 \\5.19 \end{tabular}&
\begin{tabular}[c]{@{}l@{}} 6.88\\  6.32\\  5.81 \\5.41 \\4.93 \end{tabular}
\\ \cline{2-9} 
 & \begin{tabular}[c]{@{}l@{}}mnv2\\ \textbf{mnv2\textsubscript{KD}} \\ efn \end{tabular} 
 & \begin{tabular}[c]{@{}l@{}} 9.07  \\ 8.57    \\   7.86 \end{tabular}
 & \begin{tabular}[c]{@{}l@{}} 16.06 \\ 15.06   \\   14.15 \end{tabular}                                        
 & \begin{tabular}[c]{@{}l@{}} 9.16  \\ 8.81    \\   7.97 \end{tabular}
 & \begin{tabular}[c]{@{}l@{}} 8.72  \\ 8.15    \\   7.60 \end{tabular}                                            
 & \begin{tabular}[c]{@{}l@{}} 9.11  \\ 8.75    \\   8.43 \end{tabular}                                            
 & \begin{tabular}[c]{@{}l@{}} 10.46 \\ 9.92    \\   9.28 \end{tabular}                                            
 & \begin{tabular}[c]{@{}l@{}} 9.88  \\ 9.40    \\   8.69 \end{tabular}                                            \\ \hline

\multirow{2}{*}{FR}
& \begin{tabular}[c]{@{}l@{}}DVLN~\cite{wu2017leveraging}\\ LAB~\cite{wu2018look}\\ ResNet50(Wing+PDB)~\cite{feng2018wing} \\3DDE~\cite{valle2019face}   \\GV(HRNet)~\cite{xiong2020gaussian}
\end{tabular}&
\begin{tabular}[c]{@{}l@{}} 10.84\\ 7.56 \\ 6.00 \\ 5.04 \\ 3.52\end{tabular}&
\begin{tabular}[c]{@{}l@{}} 46.93\\ 28.83\\ 22.70\\ 22.39\\ 16.26\end{tabular}&
\begin{tabular}[c]{@{}l@{}} 11.15\\ 6.37 \\ 4.78 \\ 5.41 \\ 2.55\end{tabular}&
\begin{tabular}[c]{@{}l@{}} 7.31 \\ 6.73 \\ 4.30 \\ 3.86 \\ 3.30\end{tabular}&
\begin{tabular}[c]{@{}l@{}} 11.65\\ 7.77 \\ 7.77 \\ 6.79 \\ 3.40\end{tabular}&
\begin{tabular}[c]{@{}l@{}} 16.30\\ 13.72\\ 12.50\\ 9.37 \\ 6.79\end{tabular}&
\begin{tabular}[c]{@{}l@{}} 13.71\\ 10.74\\ 7.76 \\ 6.72 \\ 5.05\end{tabular}
\\ \cline{2-9} 
 & \begin{tabular}[c]{@{}l@{}}mnv2\\ \textbf{mnv2\textsubscript{KD} }\\ efn \end{tabular} 
 & \begin{tabular}[c]{@{}l@{}}  27.12  \\ 24.08  \\  19.68  \end{tabular}                                          
 & \begin{tabular}[c]{@{}l@{}}  86.50  \\ 81.59  \\  70.55 \end{tabular}                                    
 & \begin{tabular}[c]{@{}l@{}}  27.70  \\ 27.38  \\  19.42 \end{tabular}                                             
 & \begin{tabular}[c]{@{}l@{}}  23.35  \\ 19.91  \\  16.04 \end{tabular}                                            
 & \begin{tabular}[c]{@{}l@{}}  25.24  \\ 22.33  \\  23.30 \end{tabular}                                            
 & \begin{tabular}[c]{@{}l@{}}  36.27  \\ 33.42  \\  29.21 \end{tabular}                                    
 & \begin{tabular}[c]{@{}l@{}}  33.63  \\ 29.10  \\  23.80 \end{tabular}
 \\ \hline
\multirow{2}{*}{AUC}
& \begin{tabular}[c]{@{}l@{}}
DVLN~\cite{wu2017leveraging}\\ LAB~\cite{wu2018look}\\ ResNet50(Wing+PDB)~\cite{feng2018wing} \\3DDE~\cite{valle2019face}  \\GV(HRNet)~\cite{xiong2020gaussian}
\end{tabular} &
\begin{tabular}[c]{@{}l@{}} 0.4551 \\ 0.5323 \\ 0.5504 \\ 0.5544 \\0.5775 \end{tabular} &
\begin{tabular}[c]{@{}l@{}} 0.1474 \\ 0.2345 \\ 0.3100 \\ 0.2640 \\0.3166 \end{tabular} &
\begin{tabular}[c]{@{}l@{}} 0.3889 \\ 0.4951 \\ 0.4959 \\ 0.5175 \\0.5636 \end{tabular} &
\begin{tabular}[c]{@{}l@{}} 0.4743 \\ 0.5433 \\ 0.5408 \\ 0.5602 \\0.5863 \end{tabular} &
\begin{tabular}[c]{@{}l@{}} 0.4494 \\ 0.5394 \\ 0.5582 \\ 0.5536 \\0.5881 \end{tabular} &
\begin{tabular}[c]{@{}l@{}} 0.3794 \\ 0.4490 \\ 0.4885 \\ 0.4692 \\0.5035 \end{tabular} &
\begin{tabular}[c]{@{}l@{}} 0.3973 \\ 0.4630 \\ 0.4918 \\ 0.4957 \\0.5242 \end{tabular} 
\\ \cline{2-9} 
& \begin{tabular}[c]{@{}l@{}}   mnv2   \\ \textbf{mnv2\textsubscript{KD}} \\efn\end{tabular}         
&\begin{tabular}[c]{@{}l@{}}    0.3758\\  0.4134    \\    0.4755   \end{tabular}
& \begin{tabular}[c]{@{}l@{}}   0.0321\\  0.0530    \\    0.0905\end{tabular}
&\begin{tabular}[c]{@{}l@{}}    0.3200\\  0.3570    \\    0.4366  \end{tabular}
&\begin{tabular}[c]{@{}l@{}}    0.4015\\  0.4377    \\    0.4944  \end{tabular}
& \begin{tabular}[c]{@{}l@{}}   0.3651\\  0.3940    \\    0.4493  \end{tabular}
&\begin{tabular}[c]{@{}l@{}}    0.3092\\  0.3390    \\    0.3842  \end{tabular}
&\begin{tabular}[c]{@{}l@{}}    0.3166\\  0.3521    \\    0.4067  \end{tabular}                                                                    
\\ \hline
\end{tabular}}
\end{table*}

\subsection{Comparison with Existing Approaches} \label{sec:result_comparison}
In order to assess the performance of our proposed KD architecture, we conducted three different experiments. We report the performance of our Tough-Teacher, which is EfficientNet-B3~\cite{tan2019efficientnet}, named \textit{efn}, our Student-Network, which is MobileNetV2~\cite{sandler2018MobileNetV2}, named as \textit{mnv2\textsubscript{KD}}, as well as MobileNetV2~\cite{sandler2018MobileNetV2}, named \textit{mnv2}.

\subsubsection{Evaluation on COFW}
Table~\ref{tbl:tbl_results_cofw} shows the state-of-the-art results as well as our Teacher and Student networks. As shown, Student-Network, called mnv2\textsubscript{KD}, achieves 4.11\% NME with 2.36\% FR, while the these metrics are 5.04\% and 3.74\% for MobileNetV2~\cite{sandler2018MobileNetV2} respectively. The table shows that training our proposed \textit{Student-Teacher} architecture results in significantly better performance in comparison to the base-network, MobileNetV2~\cite{sandler2018MobileNetV2}. 

\subsubsection{Evaluation on 300W}
Table~\ref{tbl:tbl_results_300w} shows a comparison between mnv2\textsubscript{KD}, mnv2 and the state-of-the-art methods on 300W~\cite{sagonas2013300} dataset. While the calculated NME for the \textit{Teacher network}, and MobileNetV2~\cite{sandler2018MobileNetV2} over the \textit{Challenging} set are 5.80\% and 6.84\% respectively, mnv2\textsubscript{KD} achieves 6.13\% by far outperforms MobileNetV2~\cite{sandler2018MobileNetV2}. For the \textit{Common} set, the NME for Student network is 3.56\%, which is better than mnv2 NME, 3.93\%. The reduction in NME for the \textit{Challenging} subset, which is about 0.71~\%, is much higher than the Common subset (about 0.37\%), showing that the proposed KDLoss function performs much better on the challenging faces. The calculated NME for the \textit{Full} subset are 4.06\% as well 4.50\% for Student network and mnv2 respectively, indicating about 0.44\% reduction. The results show that the proposed KDLoss performs a vital role in better training of the \textit{Student-Model} leading to better performance.

\begin{figure*}[t]
  \centering
  \includegraphics[width=18.5cm]{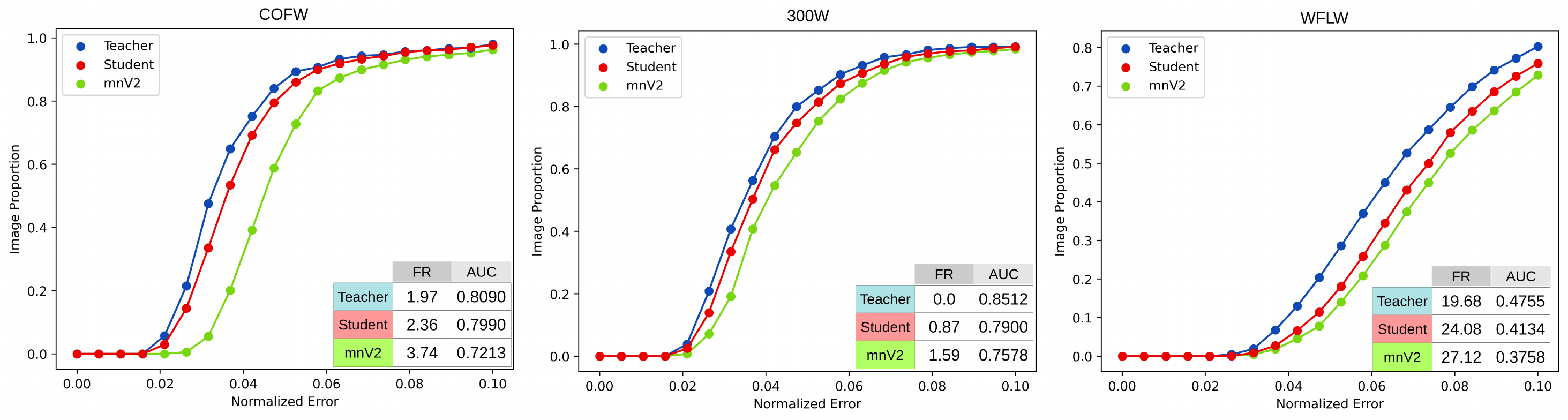}
  \caption{CED curve, FR (in \%), and AUC  being generated using \textit{Teacher network}, Student-Network, and MobileNetV2~\cite{sandler2018MobileNetV2} on COFW~\cite{burgos2013robust}, 300W~\cite{sagonas2013300}, and WFLW~\cite{wu2018look}.}
  \label{fig:CED_all}
\end{figure*}

\subsubsection{Evaluation on WFLW}
In Table~\ref{tbl:tbl_results_wflw} we compare the performance of the recently proposed methods as well as our Student-Network, mnv2\textsubscript{KD}, \textit{Teacher network} and MobileNetV2~\cite{sandler2018MobileNetV2} on WFLW~\cite{wu2018look} and its 6 subsets. Although the performance of mnv2\textsubscript{KD} does not outperform the state-of-the-art methods, the performance of mnv2\textsubscript{KD} is better than mnv2 in terms of NME (from 9.07\% reduced to 8.57\%), FR (from 27.12\% reduced to 24.08\%), and AUC (from 0.3758 increased to 0.4134). Similar to evaluation on 300W~\cite{sagonas2013300} dataset, mnv2\textsubscript{KD} performs much better over the pose set, faced 1.0\%(from 16.06\% to 15.06\%), 4.91\% (from 86.50\% to 81.59\%) reduction on NME, and FR respectively, as well as 2.09\% increase (from 0.0321 to 0.0530) on AUC. In addition, we discovered that MobileNetV2~\cite{sandler2018MobileNetV2} is not able to perform very well over this dataset, as the number of its parameters (about 2.4M) might not be enough for predicting 98 different pair of landmark points. However, the qualitative study in Sec.~\ref{sec:qualitative} shows that mnv2\textsubscript{KD} has an acceptable qualitative performance.

\begin{figure*}[t]
  \centering
      \includegraphics[width=18.5cm]{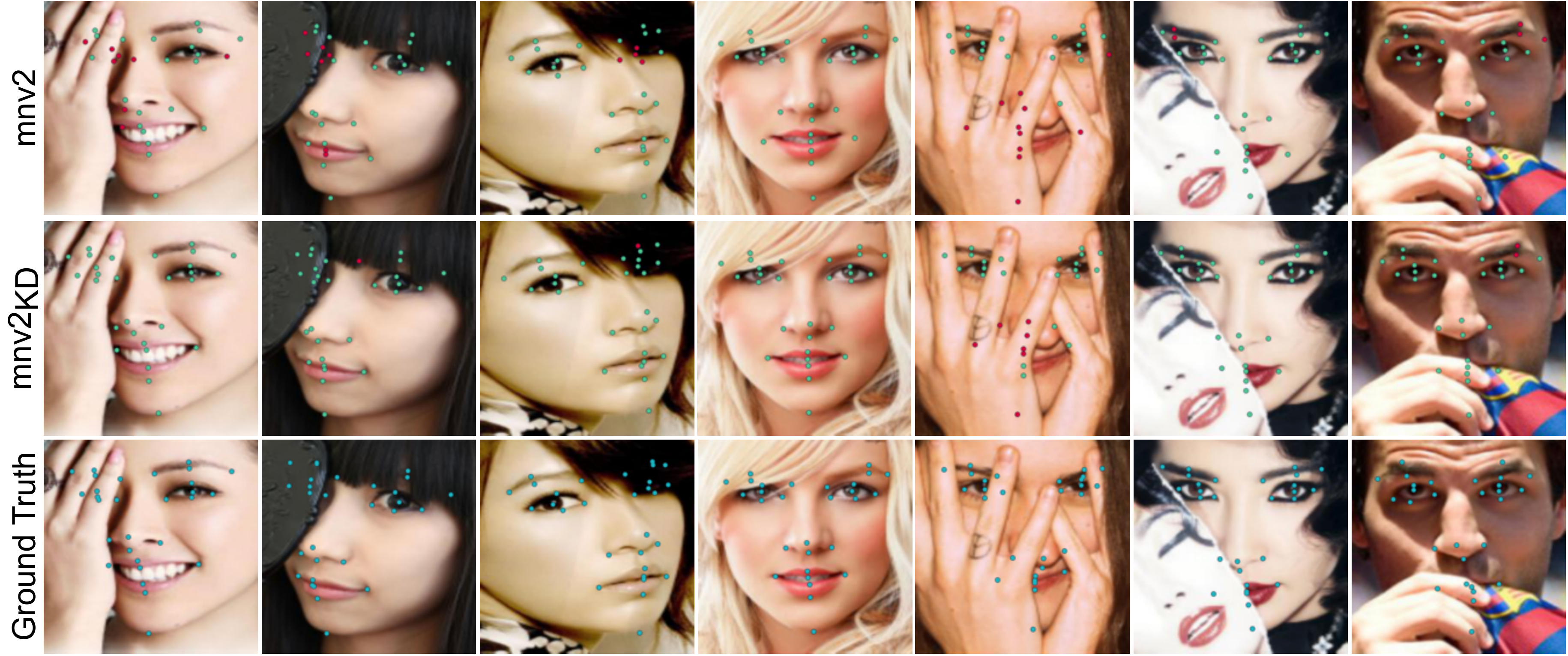}
  \caption{Face alignment using mnv2\textsubscript{KD} and mnv2 on COFW~\cite{burgos2013robust} dataset. For each landmark point if the error rate with respect to the normalization factor, is more than $0.1$, it is considered as a failure and we printed it red, and otherwise it is green.}
  \label{fig:result_Cofw}
\end{figure*}
% 

% ==================================
% =====================================
\subsection{Cumulative Error Distribution Curve Comparison}
Cumulative Error Distribution (CED) curve is the cumulative distribution function of the normalized mean error(NME). Although NME can be considered as a good evaluation metric, it is very sensitive to the outliers. As such, when the average error over the testing set is low except for some outlier samples, NME can dramatically be aggravated.

Hence, we visualize the CED curve with the failure rate threshold as \textit{0.1} for \textit{Teacher network}, Student network and the base-network, MobileNetV2~\cite{sandler2018MobileNetV2}. Fig.~\ref{fig:CED_all} shows the CED curves for COFW~\cite{burgos2013robust}, the \textit{Full} subset of 300W~\cite{sagonas2013300}, and the \textit{Full} subset of WFLW~\cite{wu2018look}. We also show FR (in \%) and AUC to better compare performance of our proposed KD-Loss as well as the KD-based architecture. As shown in Fig.~\ref{fig:CED_all}, the performance of Student network is much better than the MobileNetV2~\cite{sandler2018MobileNetV2} and a bit less accurate than \textit{Teacher network} on COFW~\cite{burgos2013robust} dataset . Similarly, on 300W~\cite{sagonas2013300}, Student network performs better than MobileNetV2~\cite{sandler2018MobileNetV2}, but its performance is not as good as the \textit{Teacher network}. On WFLW~\cite{wu2018look}, although the performance of Student network is better than MobileNetV2~\cite{sandler2018MobileNetV2}, \textit{Teacher network} performs much better than both former networks.

According to Fig.~\ref{fig:CED_all}, our proposed Student network performs much better comparing to the base-network on all datasets. However, as the the CED curves which are depicted in Fig.~\ref{fig:CED_all} show, the best accuracy improvement accrues on COFW~\cite{burgos2013robust} (the CED Student network curve is much closer to the \textit{Teacher network} curves). We conclude that since the number of parameters in our base-network, MobileNetV2~\cite{sandler2018MobileNetV2} (about 2.44 million), is much smaller than that of \textit{Teacher network} (about 12 million), its accuracy has a heavy reliance on the number of facial landmark points that exists in a dataset. Consequently, the best improvement in the results that Student network achieved comparing to the \textit{Teacher network} is on COFW~\cite{burgos2013robust}, containing only 29 landmark points, and the least improvement on WFLW~\cite{wu2018look}, with 98 landmark points.

% ==================================
% =====================================

\subsection{Model Size and Computational Cost Study}
To evaluate the model size and computational complexity, we report the number of network parameters and the FLOPs (over the resolution of 224 $\times$ 224 ) in Table~\ref{tbl:size_cost_analysis}. According to Table~\ref{tbl:size_cost_analysis}, our Student network, mnv2\textsubscript{KD}(the architecture is same as MobileNetV2~\cite{sandler2018MobileNetV2}) has only $2.4M$ parameters as well as $0.6G$ FLOPs indicating that the models is very efficient, while the its accuracy is comparable with the state-of-the-art methods. Only ASMNet~\cite{fard2021asmnet} has smaller number of parameter (1.43M), and FLOPs (0.51G) compared to the mnv2\textsubscript{KD}, while its accuracy is much less than our Student network. 

Model size is another important factor in the context of lightweight CNNs. In Table~\ref{tbl:tbl_model_size} we compare the size of our proposed mnv2\textsubscript{KD} model with the recently proposed lightweight models for face alignment. According to Table~\ref{tbl:tbl_model_size}, the size of mnv2\textsubscript{KD} is only 3.2MB, which is the smallest among other proposed models. The model size has a heavy reliance on the time of loading the graph of a model to the memory (either GPU or RAM). Accordingly, our mnv2\textsubscript{KD} can be considered to be more efficient (in terms of the model size) than the other methods reported in Table~\ref{tbl:tbl_model_size}.

\begin{figure*}[t!]
  \centering
  \includegraphics[width=18.5cm]{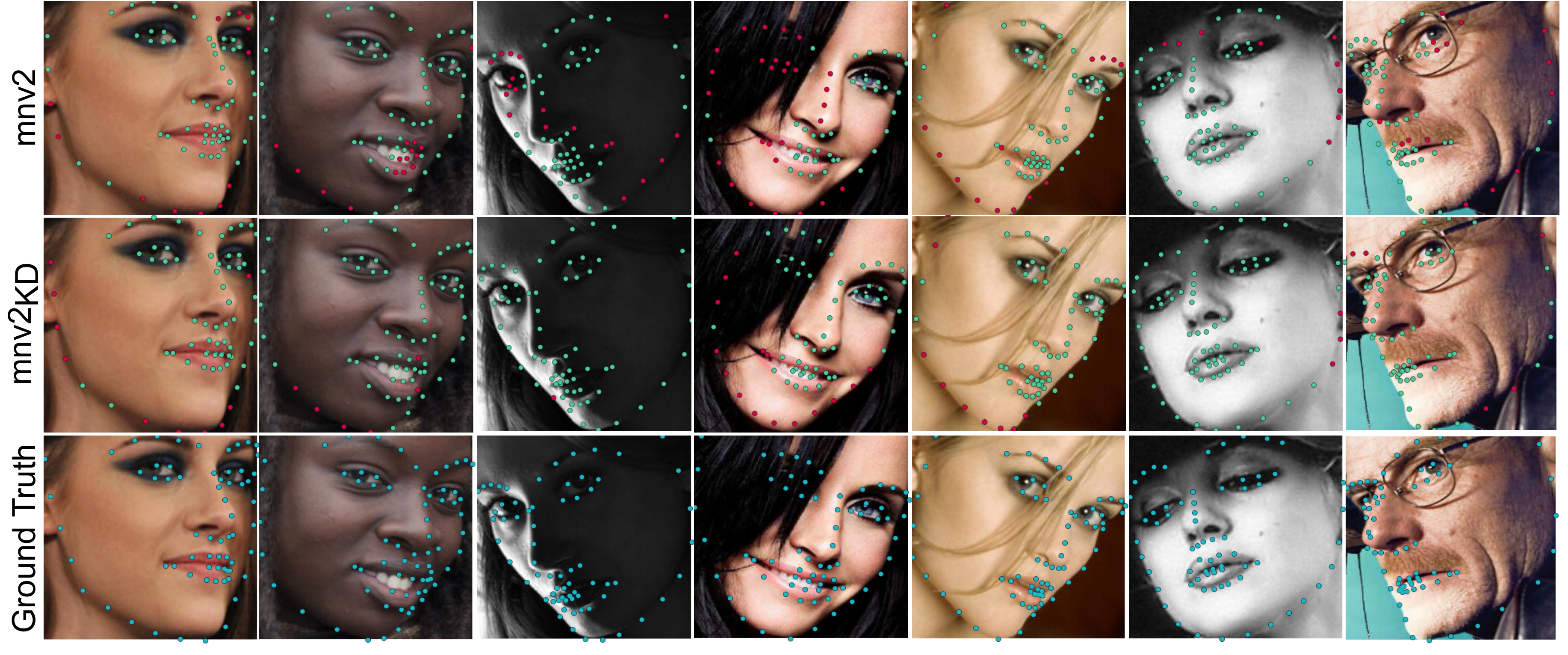}
  \caption{Face alignment using mnv2\textsubscript{KD} and mnv2 on the 300W~\cite{sagonas2013300}. For each landmark point if the error rate with respect to the normalization factor, is more than 0.1, it is considered as a failure and we printed it red, and otherwise it is green.}
  \label{fig:300W_samples_mnKD}
\end{figure*}
\begin{figure}[t]
  \centering
      \includegraphics[width=\columnwidth]{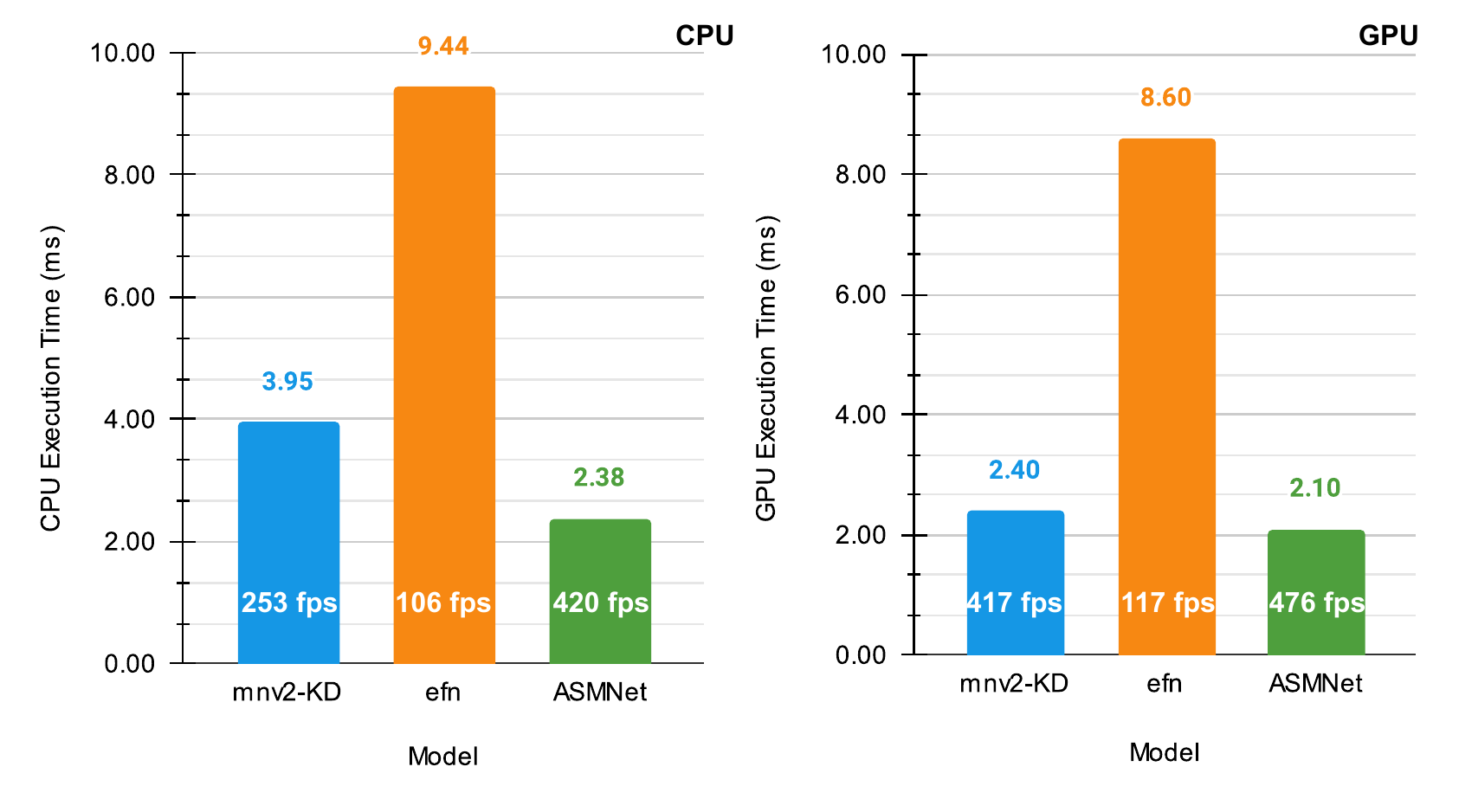}
  \caption{Comparison of the model response time and the fps of 3 lightweight models.}
  \label{fig:execution_time}
\end{figure}

As Fig.~\ref{fig:execution_time} shows, we compare the model response time as well as the fps of our mnv2\textsubscript{KD} (the Student network), efn (the Teacher network), and ASMNet. For the first experiment shown in Fig.~\ref{fig:execution_time} (the left figure), we used an Intel i7-6850K CPU and repeated each experiments 100 times and report the average response time. As Fig.~\ref{fig:execution_time} shows, on CPU, the response time of mnv2\textsubscript{KD} is 3.95ms, while the response time is 9.44ms and 2.38ms for the Teacher network and ASMNet~\cite{fard2021asmnet}, respectively. On CPU, the speed of mnv2\textsubscript{KD} is very close to the ASMNet~\cite{fard2021asmnet} (which is the smallest in terms of the number of the parameters and FLOPs among all the state-of-the-art models in the context of face alignment), but its accuracy is much better. Moreover, mnv2\textsubscript{KD} can reaches to 253 fps on CPU indicating the efficiency of the model. For the second experiment, we used a 1080ti GPU. As Fig.~\ref{fig:execution_time} shows, our mnv2\textsubscript{KD} model can reach to 417 fps indicating how fast the model can perform for detecting the landmark points in a facial image.

In addition, our proposed models do not have any post-processing costs, while for the heatmap-based models, converting heatmaps to points should be considered. By utilizing our proposed Tough and Tolerant Teachers and KD-Loss, we improve the accuracy of MobileNetV2~\cite{sandler2018MobileNetV2}, known as one of the best lightweight models, and consequently, we create a balance between efficiency and accuracy.
\begin{figure*}[t]
  \centering
  \includegraphics[width=18.5cm]{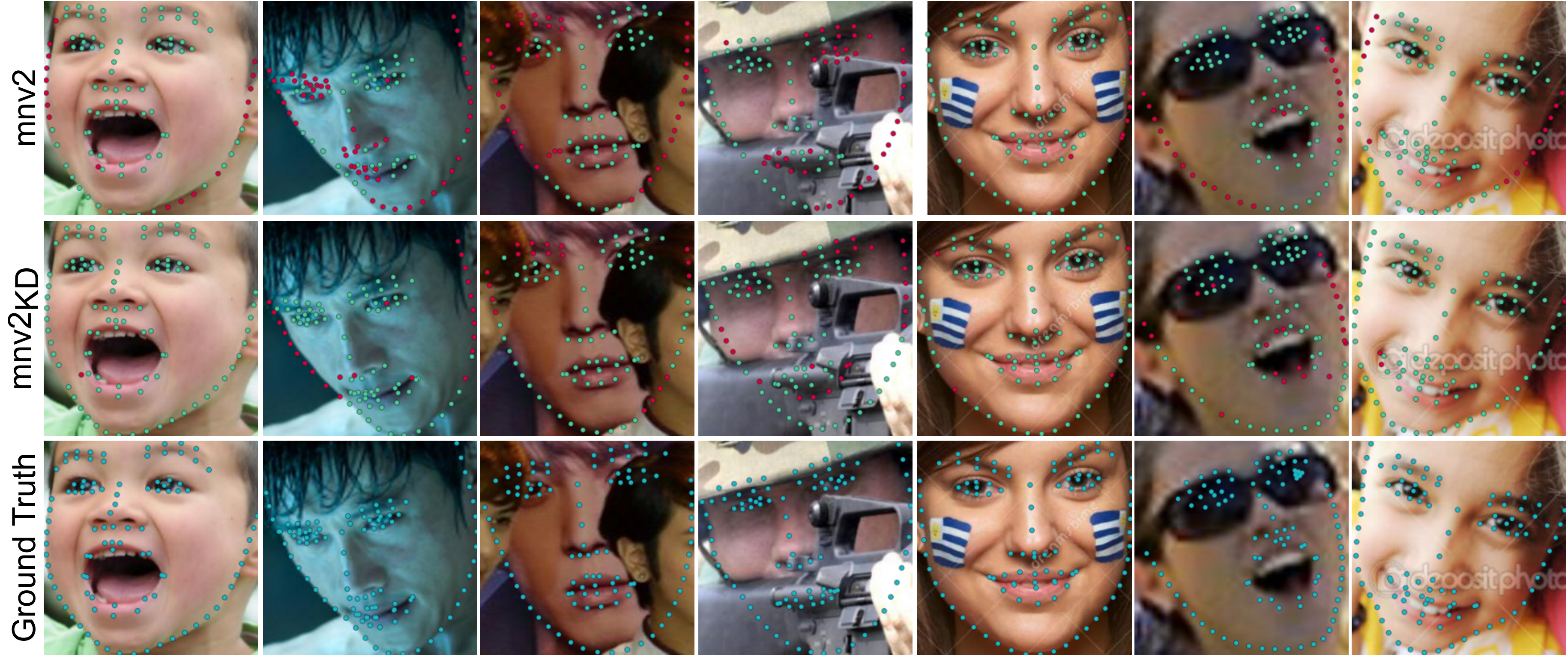}
  \caption{Face alignment using mnv2\textsubscript{KD} and mnv2 on WFLW~\cite{wu2018look} dataset. For each landmark point if the error rate with respect to the normalization factor, is more than $0.1$, it is considered as a failure and we printed it red, and otherwise it is green.}
  \label{fig:wflw_samples_mn}
\end{figure*}
% 

% ==================================
% =====================================
\subsection{Ablation Study}
In order to figure out the effectiveness of using Tough-Teacher as well as Tolerant-Teacher, we train two different model on 300W~\cite{sagonas2013300} dataset, mnv2\textsubscript{Tou} using only Accurate-landmarks, as well as mnv2\textsubscript{Tol} using only Smooth-landmarks. Then we modify the proposed KD-Loss function such that the former only defined using $ALoss_{Tou}$ and the latter using $ALoss_{Tol}$. As it is shown in Table.~\ref{tbl:study-kd-effect}, although using the modified KD-Loss results in better performance compared to the original mnv2 being trained using L2 loss, mnv2\textsubscript{Tou} performs worse than mnv2\textsubscript{Tol}. It indicates that since the distribution of Accurate-landmarks is much harder for the lightweight MobileNetV2~\cite{sandler2018MobileNetV2} to be learned in comparison to the distribution of Smooth-landmarks, the modified KD-Loss generated using only Accurate-landmarks performs not much better when compared to the original L2 loss. The better performance of mnv2\textsubscript{Tol} highlights the effect of different regions that are defined using the assistant loss functions. In KD-Loss, since the distribution of the Smooth-landmarks is much easier than distribution of the Accurate-landmarks, the \textit{Negative Assistant} region defined by $ALoss_{Tol}$ is able to guide the Student network towards the Accurate-landmarks. Accordingly, when we define the KD-Loss using only $ALoss_{Tou}$ the performance of the corresponding mnv2\textsubscript{Tou} is not as good as the performance of mnv2\textsubscript{Tol}.

\begin{table}[t!]
\caption{ Comparison of different methods in terms of number of parameters and Flops.}
\centering
\small
\label{tbl:size_cost_analysis}
\resizebox{\columnwidth}{!}{
\begin{tabular}{l c c c}

\hline
Method & Backbone & \multicolumn{1}{c}{\#Params (M)} & \multicolumn{1}{c}{FLOPs (B)}  \\ \hline \hline 
DVLN~\cite{wu2017leveraging}                    & VGG-16                & 132.0     & 14.4  \\ 
SAN~\cite{dong2018style}                        & ResNet-152            & 57.4      & 10.7 \\ 
LAB~\cite{wu2018look}                           & Hourglass             & 25.1      & 19.1 \\ 
ResNet50 (Wing + PDB)~\cite{feng2018wing}       & ResNet-50             & 25        & 3.8  \\ 
HRNetV~\cite{sun2019high}                       & HRNetV2-W18           & 9.3       & 4.3  \\ 
ASMNet~\cite{fard2021asmnet}                    &reduced MobileNetV2    & 1.43      & 0.51 \\  
EfficientNet-B3 ~\cite{tan2019efficientnet}     &   -                   & 12        & 1.8  \\ 
\textbf{mnv2\textsubscript{KD}}                & MobileNetV2            & 2.4       & 0.6  \\ \hline

\end{tabular}}
\end{table}
\begin{table}[t!] 
\caption{Comparison of the model size of the lightweight models for face alignment.}
\label{tbl:tbl_model_size}
\centering
\small
\resizebox{\columnwidth}{!}{
\begin{tabular}{p{1.5cm}p{0.5cm}p{0.9cm}p{1.5cm}p{1.8cm}p{2.0cm}}
\hline
Method          &  efn 
                & \textbf{mnv2\textsubscript{KD}}
                & ASMNet \cite{fard2021asmnet}
                &MuSiCa68 \cite{shapira2021knowing}
                & G\&LSR$\omega$\cite{shao2021robust}
\\  \hline \hline

Model Size (MB) & \multicolumn{1}{c}{7.7}
                &\multicolumn{1}{c}{3.2} 
                &\multicolumn{1}{c}{3.6} 
                &\multicolumn{1}{c} {9.1}
                &\multicolumn{1}{c} {5.9}
\\
\hline
\end{tabular}
}
\end{table}

Besides, although mnv2\textsubscript{Tol} performs better than mnv2\textsubscript{Tou}, it cannot outperform the performance of mnv2\textsubscript{KD} since Smooth-landmarks is not accurate enough to be used as the \textit{only} Teacher. Using both Teachers simultaneity with the conjunction of the proposed $ALoss_{Tou}$ and $ALoss_{Tol}$ achieves the best accuracy, as Tolerant-Teacher guides the Student network towards learning the Accurate-landmarks and Tough-Teacher it the network towards learning the original ground truths.

We compare our KD-Loss with the other widely-used loss functions for face alignment in Table~\ref{tbl:tbl_different_loss_affect}. In Table~\ref{tbl:tbl_different_loss_affect}, we also present the effect of different loss functions by comparing the NME of the face alignment task using MobileNetV2~\cite{sandler2018MobileNetV2} as the network on 300W~\cite{sagonas2013300} dataset. On all the 3 subsets, the L2 loss performs the least accurate face alignment, while KD-Loss achieves the best accuracy. In addition, the NME generated by the L1 Loss and the Smooth L1 loss are very similar. On the \textit{Challenging} subset of 300W~\cite{sagonas2013300} dataset, the NME reduces from 6.84\% for L2 loss to 6.33\% for Smooth L1 loss (0.51\% reduction), and then to 6.13\% for KD-Loss (0.2\% reduction compared to Smooth L1). For the \textit{Common} subset of 300W~\cite{sagonas2013300} dataset, the NME of the MobileNetV2~\cite{sandler2018MobileNetV2} trained using the L1 loss is 3.93\%, which reduces to 3.66\% using the Smooth L1 and then reduces to 6.13\% for training the model using KD-Loss. Hence, KD-Loss performers more accurately compared to other widely-used loss functions.

\begin{table}[t]
\caption{Comparing the performance of mnv2\textsubscript{KD}, mnv2\textsubscript{Tou}, mnV2\textsubscript{Tol}, and mnV2 with respect to NME (in \%), FR (in \%), and AUC on 300W~\cite{sagonas2013300}.}
\label{tbl:study-kd-effect}
\centering
\small
% \resizebox{\columnwidth}{!}{
\begin{tabular}{lccccc}
\hline 
\multicolumn{2}{l}{}             & mnv2   & mnv2\textsubscript{Tou} & mnv2\textsubscript{Tol} & \textbf{mnv2\textsubscript{KD}}  \\ \hline \hline

\multirow{3}{*}{NME} & Challenging & 6.84   & 6.59           & 6.41         & 6.13   \\ 
                     & Common      & 3.93   & 3.85           & 3.73         & 3.56   \\ 
                     & Full        & 4.50   & 4.39           & 4.25         & 4.06   \\  \hline
\multirow{3}{*}{FR}  & Challenging & 7.40   & 5.92           & 4.44         & 3.70   \\ 
                     & Common      & 0.18   & 0.18           & 0.18         & 0.18   \\ 
                     & Full        & 1.59   & 1.30           & 1.01         & 0.87   \\  \hline
\multirow{3}{*}{AUC} & Challenging & 0.5425 & 0.5509         & 0.5736       & 0.6029 \\ 
                     & Common      & 0.8102 & 0.8146         & 0.8230       & 0.8356 \\ 
                     & Full        & 0.7578 & 0.7630         & 0.7742       & 0.7900 \\ 
                     
\hline
\end{tabular}
% }
\end{table}

\begin{table}[t!] 
\caption{NME (in \%) of different loss functions using MobileNetV2~\cite{sandler2018MobileNetV2} for 68-point landmarks detection on 300W~\cite{sagonas2013300}.}

\label{tbl:tbl_different_loss_affect}
\centering
\small
\resizebox{\columnwidth}{!}{\begin{tabular}{l c c c}
\hline 

Loss Function                       & Common & Challenging  & Fullset        \\ \hline \hline

L2 Loss                                     & 3.93      & 6.84              & 4.50          \\
L1 Loss                                     & 3.67      & 6.38              & 4.20          \\
smooth L1~\cite{rashid2017interspecies}     & 3.66      & 6.33              & 4.18          \\
% ASMLoss~\cite{fard2021asmnet}               & 3.88      & 7.35          & 4.59              \\ 
\textbf{KD Loss}                            & 3.56     & 6.13       & 4.06                  \\ 

\hline
\end{tabular}}
\end{table}

\subsection{Qualitative Results} \label{sec:qualitative}
To better shows the quality of our proposed KD-based architecture and loss function, we provide some examples of facial landmark detection using our Student network, and the base-network, MobileNetV2~\cite{sandler2018MobileNetV2}. The result shows that Student network outperforms the base-network. Fig.~\ref{fig:result_Cofw} shows some example of face alignment using on COFW~\cite{burgos2013robust} dataset. In addition, Fig.~\ref{fig:wflw_samples_mn} shows examples of facial landmark detection on WFLW~\cite{wu2018look} dataset. Finally, in Fig.~\ref{fig:300W_samples_mnKD} we display examples of face alignment on 300W~\cite{sagonas2013300} dataset.

%-------------------SEC: CONCLUSION------------------------------------------------------
\section{Conclusion and Future Work}
\label{sec:conclusion}
This paper proposed a novel architecture inspired by the KD concept for the facial landmark detection task. Using the Active Shape Model~\cite{cootes2000introduction}, we defined Mean-landmark, Soft-landmarks as well as Hard-landmarks terms, and, used them to train our proposed Tolerant as well as Tough Teacher networks, which are EfficientNet-B3~\cite{tan2019efficientnet}. In addition, we used MobileNetV2~\cite{sandler2018MobileNetV2} as our Student-Network. The main novelty and idea of our paper are to design KD-Loss as well as a \textit{Teacher-Student} architecture utilizing a Tough and a Tolerant Teacher to help a lightweight Student network to learn the facial landmark points better. Moreover, we proposed KD-Loss, an adaptive point-wise loss function that uses the corresponding landmark points to direct the network toward the ground truth. The results of evaluating our proposed Student-Teacher architecture on widely used 300W~\cite{sagonas2013300}, COFW~\cite{burgos2013robust}, and WFLW~\cite{wu2018look} datasets show that the accuracy of the Student network is significantly better than the original MobileNetV2~\cite{sandler2018MobileNetV2}, specifically when it comes to faces with an extreme pose. Overall, the accuracy of Student network is comparable to state-of-the-art methods in facial landmark points detection. The proposed architecture can potentially be used in other similar computer vision tasks such as body-joint tracking. We will investigate using this method for such applications as a future research direction.
%-------------------------------------------------------------------------
\section{Acknowledgements}
We wish to show our appreciation to Dr. Julia Madsen for her great help on editing this paper.

% References
\bibliographystyle{model2-names}
\bibliography{refs}

% \clearpage

% \section*{Supplementary Material}
\end{document}